\documentclass{article}

\usepackage{microtype}
\usepackage{graphicx}
\usepackage{subfigure}
\usepackage{booktabs} % for professional tables

\usepackage{url}

\usepackage{breakurl}
\usepackage[breaklinks]{hyperref}

\usepackage[nohyperref, accepted]{arxiv_icml}
\usepackage{amsmath}
\usepackage{amssymb}
\usepackage{mathtools}
\usepackage{amsthm}

\usepackage[capitalize,noabbrev]{cleveref}

\theoremstyle{plain}

\theoremstyle{definition}

\theoremstyle{remark}

\usepackage[textsize=tiny]{todonotes}

\usepackage{xspace}
\newcommand{\icl}{ICL\xspace}

\newcommand{\ciwl}{CIWL\xspace}
\newcommand{\ciwlfull}{context-constrained in-weights learning\xspace}
\newcommand{\toymodeladj}{minimal\xspace}
\newcommand{\icleval}{ICL\xspace}
\newcommand{\iwleval}{IWL\xspace}
\newcommand{\ciwleval}{CIWL\xspace}
\newcommand{\flipeval}{Flip\xspace}

\icmltitlerunning{Strategy coopetition explains the emergence and transience of in-context learning}

\begin{document}

\twocolumn[
\icmltitle{Strategy Coopetition Explains the Emergence \\\ and Transience of In-Context Learning}

\icmlsetsymbol{equal}{*}

\begin{icmlauthorlist}
\icmlauthor{Aaditya K. Singh}{gatsby}
\icmlauthor{Ted Moskovitz}{ant}
\icmlauthor{Sara Dragutinović}{oxford}
\icmlauthor{Felix Hill}{gdm}
\icmlauthor{Stephanie C. Y. Chan}{equal,gdm}
\icmlauthor{Andrew M. Saxe}{equal,gatsby}
\end{icmlauthorlist}

\icmlaffiliation{gatsby}{Gatsby Computational Neuroscience Unit, University College London}
\icmlaffiliation{oxford}{University of Oxford}
\icmlaffiliation{ant}{Anthropic AI, work completed while at the Gatsby Unit, UCL}
\icmlaffiliation{gdm}{Google DeepMind}

\icmlcorrespondingauthor{Aaditya K. Singh}{aaditya.singh.21@ucl.ac.uk}

\icmlkeywords{Mechanistic interpretability, transformers, in-context learning, Machine Learning, strategy, cooperation, competition, coopetition}

% TLDR: We find and model cooperative and competitive dynamics (termed "coopetition") that explain the emergence and subsequent transience of in-context learning.

\vskip 0.3in
]

% this must go after the closing bracket ] following \twocolumn[ ...

% This command actually creates the footnote in the first column
% listing the affiliations and the copyright notice.
% The command takes one argument, which is text to display at the start of the footnote.
% The \icmlEqualContribution command is standard text for equal contribution.
% Remove it (just {}) if you do not need this facility.

%\printAffiliationsAndNotice{}  % leave blank if no need to mention equal contribution
\printAffiliationsAndNotice{\icmlEqualContribution} % otherwise use the standard text.

\begin{abstract}
In-context learning (ICL) is a powerful ability that emerges in transformer models,
enabling them to learn from context without weight updates. 
Recent work has established emergent \icl as a \textit{transient} phenomenon that can sometimes disappear after long training times. In this work, we sought a mechanistic understanding of these transient dynamics.  
Firstly, we find that---after the disappearance of \icl---the asymptotic strategy is a remarkable hybrid between in-weights and in-context learning, which we term ``\ciwlfull'' (\ciwl). 
\ciwl is in competition with \icl, and eventually replaces it as the dominant strategy of the model (thus leading to \icl transience). However, we also find that the two competing strategies actually \textit{share} sub-circuits, which gives rise to cooperative dynamics as well. For example, in our setup, ICL is unable to emerge quickly on its own, and can only be enabled through the simultaneous slow development of asymptotic CIWL. CIWL thus both cooperates \emph{and} competes with ICL, a phenomenon we term ``strategy coopetition''.
We propose a minimal mathematical model that reproduces these key dynamics and interactions.
Informed by this model, we were able to identify a setup where \icl is truly emergent and persistent.
\end{abstract}

\begin{figure*}[t]
    \centering
    \includegraphics[width=\linewidth]{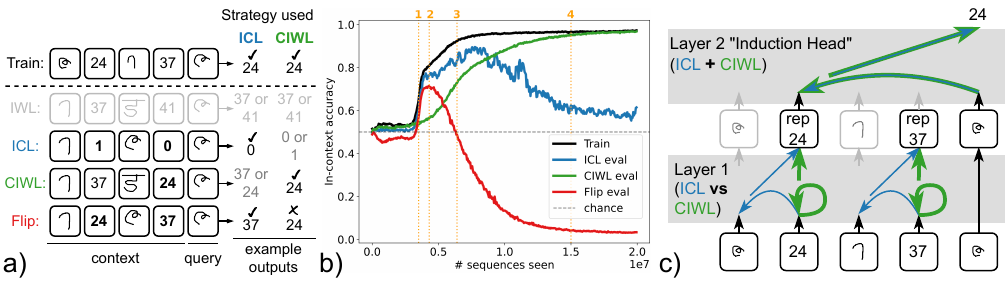}
    \vspace{-2.5em}
    \caption{\textbf{(a)} Example sequences seen during training and evaluation. Training data is ``bursty'', enabling both in-context and in-weights strategies (the context always contains an exemplar from the same class as the query, but also exemplar-label mappings are fixed throughout training). Evaluation sequences (below dotted line) are designed to measure the presence of different strategies. \icl relies on the exemplar-label mapping in context. IWL depends solely on in-weights information. CIWL requires the correct label in context, but not the query exemplar. The Flip evaluator measures the balance between \icl and \ciwl (1.0 means pure \icl, 0.0 means pure \ciwl). Bolding indicates OOD exemplar-label pairings. Grayed outputs indicate random selection between the two in-context labels. \textbf{(b)} Accuracy on sequences from (a), over the course of training. ``In-context accuracy'' is computed by restricting the network's outputs to the two labels present in context---this ensures the same chance level (0.5) for all plotted evaluators. ICL transience is clearly visible in blue. IWL is not shown, as we found little-to-no IWL in the networks (Appendix~\ref{appx:results:iwl}). We annotate four points: \textcolor{orange}{1.} the formation of Layer 2 circuits, the canonical ``induction head''; \textcolor{orange}{2.} ICL strategy dominates network output, as evidenced by peak in the Flip evaluator (red); \textcolor{orange}{3.} CIWL strategy matches strength of ICL, as indicated by 50\% performance on Flip evaluator; \textcolor{orange}{4.} CIWL strategy dominates network output, leading Flip evaluator (red) to be 0 and \ciwleval evaluator (green) to be 1. \textbf{(c)} Illustration of competitive (Layer 1) and cooperative (Layer 2) interactions we find between ICL and CIWL strategies. Both strategies are present in varying amounts through training, as represented by the varying line weights in the Layer 1 circuits: when Layer 1 acts as previous token heads, the network exhibits ICL, but when Layer 1 heads attend to self, the network exhibits CIWL. Crucially, the computation in Layer 2 remains largely unchanged after its initial formation, despite the strategy switch from ICL to CIWL.
    }
    \vspace{-1.5em}
    \label{fig:main}
\end{figure*}

\vspace{-2em}
\section{Introduction}
\label{sec:intro}

Transformer-based large language models (LLMs) show an impressive propensity for in-context learning (ICL)---the ability to use inputs at inference time to adapt behavior and solve tasks not seen in training. ICL contrasts with in-weights learning (IWL), which is standard learning through weight updates. ICL is striking not only for its power, but also because it emerges without being explicitly trained for (as \citet{GPT3} first observed, when training transformer models on internet-scale language corpora.). Newer work has shown that ICL can sometimes in fact disappear after emerging, i.e. it can be \emph{transient} over the course of training \cite{singh_transient_2023, anand_dual_2024, he_learning_2024}.

This evolving picture of ICL necessitates a deeper understanding of the dynamics of ICL emergence (and transience). ICL is often viewed to be in \textit{competition} with other strategies \cite{nguyen_differential_2024, park_competition_2024}, such as IWL, with the tradeoff thought to be modulated by data properties \cite{Chan2022, chan_toward_2024}, model size \cite{wei2023larger}, and/or training time \cite{singh_transient_2023}. While competition may explain why ICL gives way to other strategies through the course of training, the question remains: why does it emerge in the first place (if only to fade away)?

In this work, we aim to extend the \textit{mechanistic} understanding of ICL, which currently focuses on induction heads \cite{InductionHeads} and their emergence dynamics \cite{singh2024needs}, to a richer dynamical setting involving multiple strategies cycling in and out over the course of learning. 
To do so, we reproduce and investigate the key transience result in a simplified synthetic data setting with a 2-layer attention-only transformer. 
Using behavioral evaluators, we find the asymptotic strategy after the disappearance of ICL is not pure in-weights learning. Rather, it is a surprising hybrid strategy that we term \ciwlfull (\ciwl, Section~\ref{sec:ciwl}). 
The implementation of \ciwl takes the form of skip-trigrams \cite{anthropicMathFramework} distributed across multiple heads in a form of superposition \cite{elhage2022superposition}. 
Perhaps even more remarkably, we find that even though \ciwl dominates over \icl asymptotically, both strategies \textit{share} critical sub-circuits (Section~\ref{sec:whyicl:reuse}), indicating cooperative dynamics between these seemingly competitive mechanisms---a phenomenon we term ``strategy coopetition.''  We borrow the term ``coopetition'' from game theory, where it describes situations where competitors simultaneously cooperate and compete with each other.\footnote{A classic historical examples is Hollywood studios in the early 20th century, who competed aggressively for talent and audiences, but also collaborated to establish industry-wide standards and jointly negotiate with labor unions.}
Cooperation enables the emergence of \icl (despite it not being asymptotically preferred), while competition leads to its eventual transience (as previously indicated by \citet{singh_transient_2023}).
Notably, \icl emergence can only occur when \ciwl is not fully formed (Section~\ref{sec:whyicl:race}), which may explain the tradeoff seen by \citet{Chan2022}: under certain data properties, \ciwl forms too ``quickly,'' preventing \icl emergence.

We formalize our intuitions from this case study into a \toymodeladj mathematical model capturing coopetition dynamics and explaining the transience behavior. 
Our model motivates further experiments to modulate the tradeoff between strategies through learning, like reducing the asymptotic bias towards one strategy via data properties.
For the case of ICL vs CIWL, we find that matching context and query exemplars removes the asymptotic bias towards \ciwl and leads to the persistence of the ``faster'' \icl strategy.

Our work represents a step forward in understanding how different strategies trade off during learning, through mechanistic investigations on the transience of ICL. We hope our work inspires further work on such coopetitive, dynamical phenomena and enhances intuitions around how capabilities emerge (and possibly fade) when training transformers.

\section{Experimental setup}
\label{sec:setup}

\subsection{Training details}
\label{sec:setup:training}
We train 2-layer attention-only transformers \cite{vaswani2017attention,anthropicMathFramework} on a synthetic few-shot learning task.
We use $d_{model} = 64$, with $8$ heads per layer and learned absolute positional embeddings. As is common in mechanistic work \cite{InductionHeads, singh2024needs}, we chose this minimal setting as it sufficed to reproduce key phenomena.\footnote{We still consider various alternative architectures, such as those with MLPs or RoPE \cite{rope}, in Appendix \ref{appx:behavior}.} We used the Adam optimizer \cite{adam} with $\beta_1=0.9$, $\beta_2=0.999$, a learning rate of $10^{-5}$, and a batch size of 32 sequences. All models were trained in JAX \cite{jax}. All code is open-sourced at \url{https://github.com/aadityasingh/icl-dynamics}.

\subsection{Dataset}
\label{sec:setup:data}

Our few-shot learning task consists of sequences of exemplar-label pairs, where image exemplars are drawn from the Omniglot dataset of handwritten characters \cite{omniglot}. Each character class contains 20 image exemplars, and is assigned to a different one-hot label. Images were embedded using a Resnet18 encoder that was pretrained on ImageNet \cite{resnet, imagenet}, before being input to the transformer \cite{singh2024needs}. While the original Omniglot dataset has 1623 classes, we follow prior work \cite{Chan2022} and augment it to 12984 classes by applying flips and rotations. Of these, we use a random 12800 for training. In Appendix~\ref{appx:behavior:data}, we also considered using different \#'s of classes or exemplars, observing similar modulations to \citet{singh_transient_2023} for the duration, timing, and magnitude of the transience effect.

Each training sequence consists of two exemplar-label pairs (the ``context''), followed by a ``query'' exemplar. The model is tasked with outputting the correct label for the query (cross-entropy loss). We train on the \textit{``bursty''} sequences from \citet{Chan2022}, where an exemplar-label pair from the same class as the query exemplar is always present in context. This kind of burstiness reflects real-world data distributions like language, and permits both in-weights strategies (since exemplar-label mappings are fixed throughout training) and in-context strategies (since the query exemplar always comes from the same class as one of the two ``context'' exemplars) (Figure~\ref{fig:main}a, top row).

\subsection{Evaluators}
\label{sec:setup:evals}
To measure in-context and in-weights strategies, we consider four out-of-distribution evaluation sets (Figure~\ref{fig:main}a). 

In the ``\icleval'' evaluator, we replace labels from training with 0 and 1. This invalidates any exemplar-label mappings previously stored in weights. To perform above chance, the model is instead forced to use the mappings provided \textit{in context}, i.e. an \textit{in-context learning (ICL)} strategy.

In the ``\iwleval'' evaluator, the query exemplar (and its label) does not appear in the two context pairs, thus preventing the model from using context to perform the task. This forces the model to use knoweldge stored in weights, i.e. a \text{pure in-weights learning (IWL)} strategy. Performance on this evaluator barely rises above chance (Appendix~\ref{appx:results:iwl}).

The ``\ciwleval'' evaluator is like the \iwleval evaluator, except that while a matching exemplar to the query is not in context, the correct label \emph{is} in context. Following \citet{singh_transient_2023}, the correct label is randomly paired with one of the context exemplars. Like the \iwleval evaluator, this evaluator can be solved by pure IWL. However, this evaluator also permits a \emph{mixed} strategy where in-context label information can be combined with in-weights information.
We refer to the strategy that achieves above chance on this evaluator (but chance on the \iwleval evaluator) as \textit{context-constrained in-weights learning (CIWL)}. It requires the correct label token in context, but not the full example-label pairing.

Finally, the ``\flipeval'' evaluator can be seen as testing for the model's preference between \icl and \ciwl strategies. The two context exemplars have their labels flipped relative to training (e.g. if exemplars X and Y were trained with label mappings X:24 Y:25, the in-context mappings would instead be Y:25 X:24 for this evaluator). The query comes from one of those two classes. If the network prefers the ICL strategy, accuracy on this evaluator would be 1. If the network prefers \ciwl, accuracy would be 0, as it would instead output the label that was paired with the query exemplar during training.
This evaluator is especially useful in measuring which strategy is \textit{dominant} at each point in training (even when both strategies are present in some form).

\section{Reproducing the transience of emergent \icl and the convergence to \ciwl}
\label{sec:reprod}

Figure~\ref{fig:main}b shows a reproduction of the key transience phenomena in our simplified setting, with an extended figure in the appendix (Figure~\ref{fig:extended-main}). \icl emerges and then disappears, as evidenced by above-chance performance on the \icleval evaluator (blue). The disappearance of \icl corresponds to a rise in accuracy on the \ciwleval evaluator (green), indicating that the network is somehow using \textit{just the label information} from context, in combination with some form of in-weights information, to get the right answer. 

Notably, there is a significant period of time where \icl and \ciwl co-exist (from $\sim$3e6 to $\sim$2e7 sequences seen). During this time, the balance between the two strategies shifts from \icl to \ciwl, as evidenced by the decrease in the \flipeval evaluator (red). Early in training, \icl dominates (annot. \textcolor{orange}{2}). Then, \icl and \ciwl are roughly balanced (annot. \textcolor{orange}{3}), before ICL fades and CIWL dominates (annot. \textcolor{orange}{4}). This corresponds to a switch in the network's behavior, which first bases its output on the exemplar-label mappings from context, and then eventually to the exemplar-label mappings from training (in combination with the label provided in context).

\section{The asymptotic ``\ciwlfull'' (\ciwl) mechanism}
\label{sec:ciwl}

Prior work \citep[e.g.,][]{Chan2022, chan_toward_2024, nguyen_differential_2024, anand_dual_2024} has contrasted ICL to ``pure'' IWL strategies that are completely independent of context. However, we find that the dominant asymptotic mechanism in our networks is dependent on in-weights information but also \emph{context-constrained}, requiring the presence of the correct label in context. 
In this section, we thus focus on the mechanisms underlying the CIWL strategy, and relegate discussion of auxiliary pure in-weights strategies to Appendix~\ref{appx:results:iwl}.

\begin{figure}
    \centering
    \includegraphics[width=0.97\linewidth]{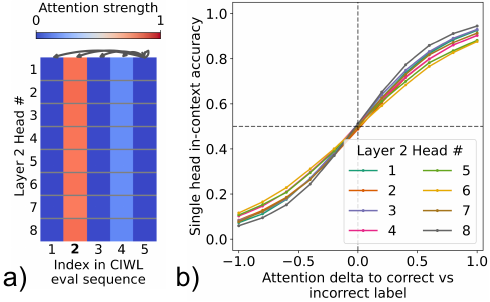}
    \vspace{-1em}
    \caption{CIWL strategy is implemented via skip-trigram-like mechanisms in Layer 2 (L2), with substantial K- and V- composition to Layer 1 (L1). \textbf{(a)} Average attention patterns for L2 heads at the end of training. Attention is measured from the query token (index 5) to each token in context. It is computed over \ciwleval sequences where the correct label is at \textbf{index 2} (results for label at index 4 in Fig~\ref{fig:ciwl2}). We see that, at the end of training, L2 heads attend to the correct label, regardless of what exemplar it is paired with in context. \textbf{(b)} Task performance as a function of clamped attention delta to correct vs. incorrect label token, calculated over 5000 \ciwleval sequences, when only the given head is active. \ciwleval accuracy increases as attention to the correct label increases.}
    \vspace{-1em}
    \label{fig:ciwl}
\end{figure}

\subsection{Layer 2 heads are skip-trigram-copiers}
\label{sec:ciwl:skiptrigram}

While label copying might appear to be a separate additive mechanism, we actually find that CIWL is implemented through the use of attention heads in Layer 2 (L2), which act as skip-trigram-copiers \cite{anthropicMathFramework}. This mechanism attends to the correct label, and then copies it forward to the output: ``... [label] ... [query] $\rightarrow$ [label].'' 

Fig~\ref{fig:ciwl}a shows attention patterns on the \ciwleval evaluator. At the end of training, the model correctly attends from the query exemplar to the in-context label that was paired with the query during training (regardless of which exemplar that label currently appears with in context). We find such trends on in-distribution training sequences as well as OOD \ciwleval and \flipeval evaluators. 
Skip-trigram heads only appear in L2, with Layer 1 (L1) heads not contributing directly to the output: ablating L1 $\rightarrow$ output connections, via the pattern preserving method of \citet{singh2024needs}, results in a negligible drop in accuracy (98.7\% to 98.5\%) on the \ciwleval evaluator.

The second part of this mechanism is the copying forward of the label from context to output. We demonstrate this through the use of causal ablations: we clamp the attention pattern for a given head to give a certain weight to the ``correct'' label token and the remaining weight to the ``incorrect'' label token. When considering each L2 head in isolation,\footnote{This is performed by taking the output of a single L2 head after our clamp, and feeding it through the rest of the network.} we find that modulating this weight is nearly perfectly correlated with output predictions (Fig~\ref{fig:ciwl}b), causally indicating that each L2 head is serving to copy input labels to the output. The combination of attending to the correct label and copying it forward give rise to the CIWL mechanism.

\textbf{Layer 2 skip-trigrams may be represented in superposition.} While the averaged results in Fig~\ref{fig:ciwl} may indicate that heads are performing similar functions to each other (in contrast to \citet{singh2024needs}, who found differing strengths for different induction heads), we find more heterogeneity when inspecting responses on individual data points (Fig~\ref{fig:superposition}a). It's also surprising that heads do not perfect their attention pattern to the correct label (again differing from the induction heads in \citet{singh2024needs}). We suspect that the large number of classes present (12800), compared to the dimension of the model (64) and number of heads (8) lead to the individual skip-trigrams (for each exemplar-label pair) being stored in superposition (similar to sparse features stored in Transformer MLPs \cite{bricken2023monosemanticity}; preliminary evidence in Appendix~\ref{appx:results:superposition}).

\subsection{Layer 1 engages in K- and V- composition, despite not being necessary}
\label{sec:ciwl:kvcomp}

While L1 heads do not directly influence network output, ablating them completely (without preserving patterns and values in L2) leads to a big drop in performance. This indicates substantial composition \cite{anthropicMathFramework} between the two layers (i.e. L2 attention heads are dependent on inputs from L1), despite the fact that skip-trigrams can be implemented with just a single layer: Appendix Fig~\ref{fig:ciwl1L} shows that a 1-layer model can learn \ciwl using skip-trigram-like circuits, in line with prior work that describes 1-layer mechanisms for skip-trigrams \cite{anthropicMathFramework}.

To dig into the specific composition between the two layers, we consider various ablations of L1 head outputs on the \ciwleval evaluator data, while preserving some combination of values, keys, or queries in L2. Recall that accuracy at the end of training on this evaluator is 98.7\%, with chance level being 50\%. When we preserve everything but values, performance drops to 59.3\%, indicating L1 outputs influence L2 values in a crucial way (known as V-composition).\footnote{As defined by \citet{anthropicMathFramework}, the degree of k-, q-, and v- composition is the degree that a previous layer contributes to keys, queries, and values (respectively) of a following layer.} 
When we preserve everything but keys, performance drops to 57.5\%, indicating K-composition. When we preserve everything but queries, performance drops to 95.4\%, indicating a very small amount of Q-composition (if at all). Thus, the primary reliance of L2 on L1 is in the calculation of L2 \textit{keys and values}, begging the question: Given that a one-layer network could implement skip-trigram-copiers, why does this observed K- and V- composition between layers emerge?
%We can further localize this composition to specific heads (Appendix~\ref{appx:results:localize}), but a larger question remains: Given that a one-layer network could implement skip-trigram-copiers, why does this observed K- and V- composition between layers emerge?

\begin{figure*}
    \centering
    \includegraphics[width=\linewidth]{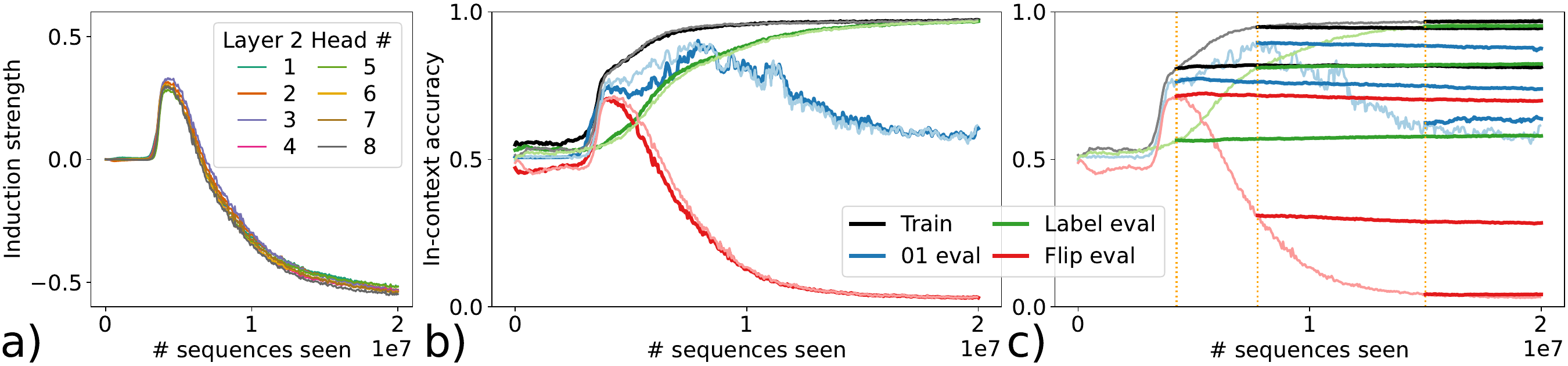}
    \vspace{-2em}
    \caption{The transience of emergent \icl is driven by changes in the function of Layer 1. \textbf{(a)} Average induction strength (attention delta to correct \icl token vs incorrect \icl token) on Flip eval data of each Layer 2 head, through the course of training. We see induction circuits emerge then flip, matching the end-of-training attention patterns shown in Fig~\ref{fig:ciwl}. \textbf{(b)} For each available checkpoint, we fix Layer 2 weights to be those from the end of training, and plot performance on each of our evaluators. Using the Layer 2 weights from the end of training (darker curves) reproduces the original behavior (lighter curves; matches Fig \ref{fig:main}b) at all points in training after the initial flat portion. This indicates that Layer 2 is not meaningfully changing during the transition from \icl to \ciwl. \textbf{(c)} For each available checkpoint, we fix the Layer 1 weights to those from a specific checkpoint (marked by the dotted orange vertical lines). After the Layer 1 weights are fixed in this way (darker curves), network behavior doesn't change, as evidenced by the flat lines on all the data we considered.}
    \vspace{-1em}
    \label{fig:l1transition}
\end{figure*}

\section{ICL emerges, despite not being asymptotic, due to cooperative interactions with CIWL}
\label{sec:whyicl}

Induction circuits (which typically underpin \icl strategies, c.f., \citet{InductionHeads}) also show characteristic K- and V- composition, and seem to be present during the transient emergence of ICL in our experiments (Fig~\ref{fig:l1transition}a). Could the K- and V- composition for the two strategies be related? Do these seemingly competing strategies actually \textit{share} circuit elements? Could that be why ICL emerges, despite not being asymptotically preferred? In this section, we find the answer is, remarkably, yes.

\subsection{The \icl to \ciwl transition is explained by L2 reuse and L1 dynamics}
\label{sec:whyicl:reuse}

Induction circuits are typically comprised of a ``previous token'' head in the 1st layer that copies information from the previous token into the next token, and an ``induction head'' in the 2nd layer that uses the 1st layer to find tokens preceded by the present token. This enables a common form of \icl: ``[A*][B*] ... [A] $\rightarrow$ [B]'' \cite{InductionHeads}. 

Surprisingly, we find here that the L2 induction heads in the \icl induction circuits are re-used by \ciwl with little change. The transition from \icl to \ciwl is implemented by a transition in the role of \textit{L1 heads}, which switch from previous-token-attention to attending-to-self (Fig~\ref{fig:main}c).\footnote{Algorithmically, we believe these heads are essentially learning to map label tokens to prototypical embeddings. It's hard to directly verify such a hypothesis, given the heads' rotation invariance and observed superposition (Appendix~\ref{appx:results:superposition}), but our evidence for the semantic stationarity of L2 points to such an algorithm.}

This is supported by attention patterns in L1 (Fig~\ref{fig:l1attn}) and L2 (Fig~\ref{fig:ciwl}a), and also by two sets of additional experiments, inspired by the notion of progress measures \cite{nanda2023progress}. We take each checkpoint from training, and then fix subsets of the network to the weights from a different checkpoint from the same run. In Fig~\ref{fig:l1transition}c, we show that fixing the first half of the network (embedding + L1) to that of a checkpoint at a given point (orange dotted lines) leads to very little change in network behavior after that point (flattening of the dark lines as compared to the lighter lines). Conjointly, if we fix the second half of the network (L2 + unembedding) to the checkpoint from the end of training (Fig~\ref{fig:l1transition}b), we see little difference in behavior after the initial phase change, which indicates that Layer 2 changes are not meaningfully affecting the network after that point. 

Notably, this means the canonical ``induction heads'' (in L2) remain, but they become part of the computational strategy \ciwl rather than \icl, due to the change in L1. 
These results connect to a broader emerging notion that few-shot ICL \cite{GPT3} may lie on a spectrum of ICL abilities \cite{lampinen2024broaderspectrumincontextlearning, yin2025attentionheadsmatterincontext}, which we now show may be connected on the mechanistic level via shared subcircuits. 

\begin{figure*}[ht]
    \centering
    \includegraphics[width=\linewidth]{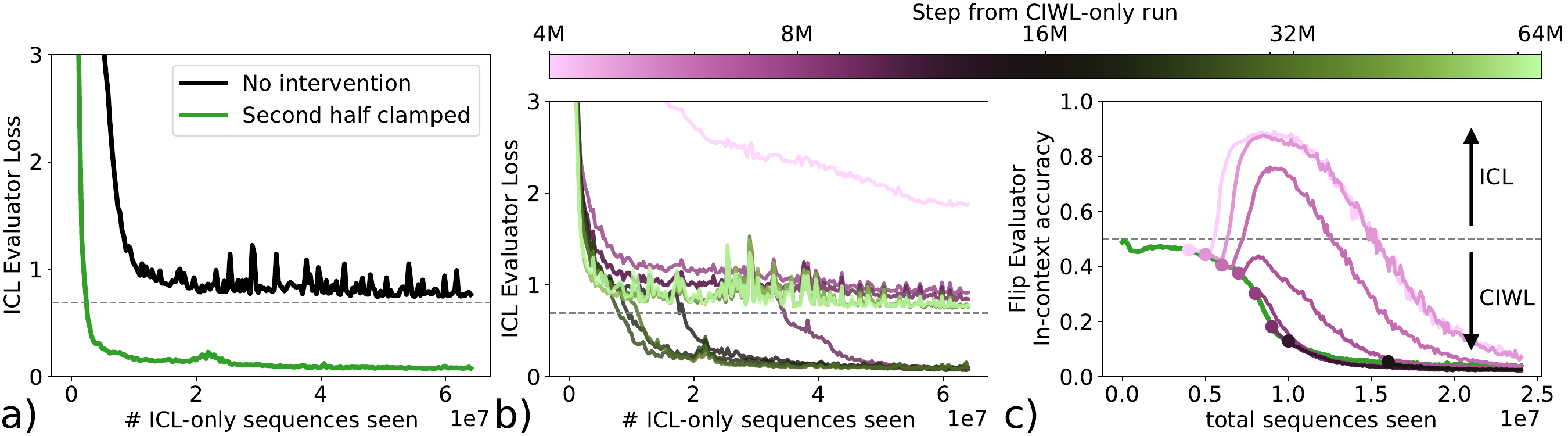}
    \vspace{-2em}
    \caption{\icl emergence is enabled due to \emph{cooperative} interactions with \ciwl. \textbf{(a)} In this plot, we train networks on ``ICL-only'' data, i.e. where ICL is a viable strategy but CIWL is not. Without any interventions (black), we hit a loss plateau that greatly slows learning \citep[c.f.,][]{singh2024needs}. However, if we replace the Layer 2 and unembedding weights with those from end-of-training on our standard training data (green), the network learns quickly. We thus see that these weights, which were part of a CIWL strategy, are reusable for learning ICL. \textbf{(b)} We further consider using Layer 2 + unembedding weights from different checkpoints of a ``CIWL-only'' run, and then training on ``ICL-only'' data. Early and late checkpoints lead to no learning of ICL, but middling checkpoints (from 9.8M to 31.5M, identified via binary search) do enable ICL learning. \textbf{(c)} We continued training different checkpoints from the ``CIWL-only'' run on our standard training data. Once CIWL has formed (later checkpoints), ICL does not re-emerge even when switching to the bursty data (which otherwise permits ICL).}
    \label{fig:cooperation}
    \vspace{-1.5em}
\end{figure*}

\subsection{Learning the \icl strategy is \textit{enhanced} by the availability of asymptotic \ciwl}
\label{sec:whyicl:iclenabled}

Why does ICL emerge at all, if it is not asymptotically preferred? We find that ICL's emergence may actually be enabled by ICL's shared L2 subcircuit with CIWL. Thus, shared subcircuits may lead to \textit{cooperative interactions}, on top of known competitive interactions \cite{singh_transient_2023, park_competition_2024}. 

To show this, we train networks on ``\icl-only'' data, where we randomize the exemplar-label mappings across contexts, but keep the mappings consistent within each context. This data can only be solved by ICL, and not by CIWL (since there are no fixed mappings to learn in weights). \footnote{This resembles the setup used by \citet{singh2024needs}, except with 12800 classes and 20 exemplars per class.} 
When training from scratch, we see the network struggle to learn (Fig~\ref{fig:cooperation}a, black). 
Yet when we previously trained on bursty data (where CIWL \emph{was} a viable strategy), we saw \icl emerge quite quickly (Fig \ref{fig:main}b). Combined, these form a striking result---ICL actually emerges more easily when it is not the only viable strategy! Presence of an asymptotic \ciwl mechanism may be enabling \icl emergence.

To further test this hypothesis, we again trained on ICL-only data, but clamped the weights of the second half of the network (L2 + unembedding), using those from the end of training from our ``standard run'' on bursty data, after the network has converged to a \ciwl strategy. Through this causal intervention on dynamics (using the framework from \citet{singh2024needs}), we show that ICL can emerge quite rapidly (Fig~\ref{fig:cooperation}a, green) with the help of the \ciwl weights for Layer 2. These findings connect to the notion that loss plateaus may arise from multiple sub-circuits needing to go right \cite{singh2024needs}. When the Layer 2 subcircuits are provided, ICL emerges quickly.

\subsection{\icl is ``close to'' the path towards \ciwl}
\label{sec:whyicl:iclclose}

One confound in the previous experiment is that \ciwl weights at the end-of-training were likely influenced by the earlier presence of \icl (a possible path dependence).
Here, we consider a stricter version of the experiment.

We start by training networks on ``\ciwl-only'' data: sequences drawn from the same distribution as the \ciwleval evaluator, which can be solved by \ciwl but not \icl. We then take checkpoints from different points in training of this \ciwl-only run, and use the L2 + unembedding weights to repeat the experiments of Section~\ref{sec:whyicl:iclenabled}; namely, clamped training on \icl-only data. Essentially, we're asking if the L2 weights when learning a \ciwl strategy, without influence from \icl, can still enable \icl strategies. Surprisingly, we find that the answer is, for a brief period, yes.

Fig~\ref{fig:cooperation}b shows training curves of our clamped networks on \icl-only data. Each curve is colored based on the iteration from the \ciwl-only run that the clamped weights come from. We see that there is a region from about 10M to 32M sequences seen where \icl can emerge. After this point, L2 weights likely specialize too strongly on a \ciwl strategy (as they're being trained on data that only permits such strategies), and thus can no longer be re-used for \icl.

\subsection{If \ciwl is fully formed, \icl will not emerge}
\label{sec:whyicl:race}

To understand why and when \icl emerges, a final critical factor is that \ciwl is ``slow'' to emerge on certain data distributions, relative to \icl.
If this were not the case, the competitive interactions might dominate and \ciwl might prevent \icl from ever emerging. Indeed, \citet{Chan2022} found many data distributions where in-weights strategies appear quickly and dominate, and in-context strategies do not emerge at all (reproduced in our setup in Appendix~\ref{appx:behavior:data}).

To show more directly that \icl cannot emerge once \ciwl has fully formed, we take various checkpoints from the ``\ciwl-only'' run, and continue training but with bursty data (Fig~\ref{fig:cooperation}c). When initializing from early ``\ciwl-only'' checkpoints, we see a transient emergence of \icl before \ciwl again dominates. At later checkpoints, \ciwl persists, with no emergence of \icl. This indicates that \icl could only emerge at the earlier training times because \ciwl was not yet fully formed.

\subsection{Strategy coopetition}
\label{sec:whyicl:coopetition}

Taken together, these results reveal a surprisingly rich interaction between \icl and \ciwl: \icl is enabled by \ciwl, yet also competing with \ciwl for resources in the network. Prior work has pointed to competition between \icl and \ciwl, which we now mechanistically observe in Layer 1. In contrast, Layer 2 exhibits cooperative interactions that lead ICL to emerge, despite not being asymptotic. This emergence dynamic has a few requirements:

1. \textbf{Useful}: \icl's ability to reduce loss on ``bursty data.''

2. \textbf{On the way}: \icl is ``close'' to the path of a 2L network learning an asymptotic \ciwl strategy.

3. \textbf{Fast}: The \ciwl strategy emerging ``slowly'' enough to allow the ``faster'' \icl to make a transient appearance.

The first factor follows from the dataset design. The second requirement is supported by Sections~\ref{sec:whyicl:reuse}-\ref{sec:whyicl:iclclose}. The third requirement is supported by Section~\ref{sec:whyicl:race} and prior work \cite{Chan2022, singh_transient_2023, nguyen_differential_2024}. 

\section{A toy model of strategy racing + coopetition}
\label{sec:toymodel}

To crystallize these intuitions, we propose a minimal mathematical model that captures the competitive and cooperative interactions at play. Specifically, we consider learning four vectors $\mathbf{a}, \mathbf{b}, \mathbf{c}, \mathbf{d}$ via gradient descent on the following loss function, where $\mathbf{a^*}, \mathbf{b^*}, \mathbf{c^*}, \mathbf{d^*}$ are the true values:
\begin{align}
    \mathcal{L}&(\mathbf{a}, \mathbf{b}, \mathbf{c}, \mathbf{d}) = \notag \\
    &\left(\underbrace{\;\left|\left|\mathbf{a^*} \otimes \mathbf{b^*} \otimes \mathbf{c^*} - \mathbf{a} \otimes \mathbf{b} \otimes \mathbf{c}\right|\right|_F^2}_{\text{Mechanism 1 (\icl) Loss}} \, + \; \mu_1\right) \label{1}\\
    &\times \left( \underbrace{\;\left|\left|\mathbf{d^*} \otimes \mathbf{b^*} \otimes \mathbf{c^*} - \mathbf{d} \otimes \mathbf{b} \otimes \mathbf{c}\right|\right|_F^2}_{\text{Mechanism 2 (\ciwl) Loss}} \right) \label{2}\\[5pt]
    &+ \alpha\underbrace{||\mathbf{a} \otimes \mathbf{d}||_F^2}_{\text{Competition}}, \label{3}
\end{align}
where $\alpha \geq 0$ is a parameter modulating the strength of competition,\footnote{An alternate interpretation is that $\alpha$ is a fixed version of a Lagrange multiplier for satisfying the constraint that $\mathbf{a}$ or $\mathbf{d}$ is 0.} and $\mu_1 \geq 0$ modulates the relative asymptotic preference of mechanism 2 over mechanism 1. If $\alpha, \mu_1 > 0$, loss is minimized when $\mathbf{a} = \mathbf{0}, \mathbf{b}=\mathbf{b^*}, \mathbf{c}=\mathbf{c^*}, \mathbf{d}=\mathbf{d^*}$.

\begin{figure}
    \centering
    \includegraphics[width=\linewidth]{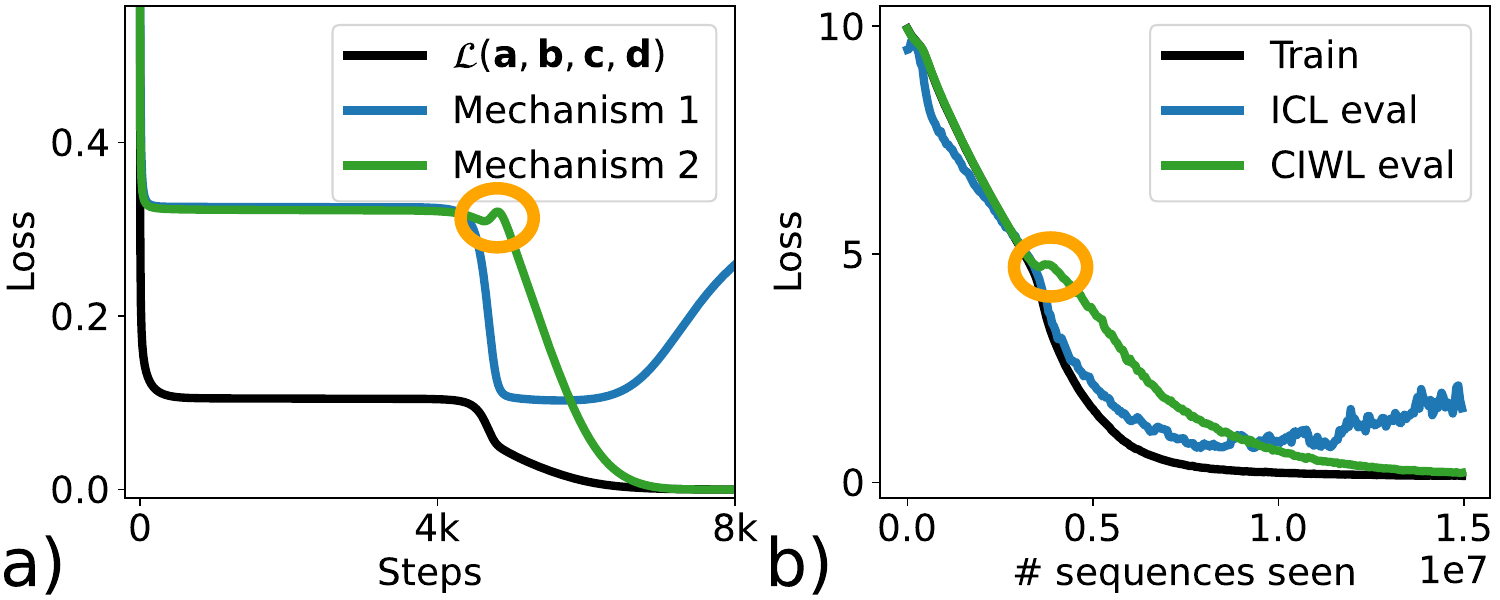}
    \vspace{-2em}
    \caption{Minimal mathematical model captures key phenomena of strategy racing and coopetition. \textbf{(a)} Toy model dynamics. \textbf{(b)} Loss of real transformer, corresponding to Fig~\ref{fig:main}b. Notably, both curves exhibit transience behavior in Mechanism 1 (ICL). Intriguingly, the toy model also captures nonmonotonicity in the emergence of Mechanism 2 (CIWL), highlighted in orange.}
    \label{fig:toymodel}
    \vspace{-1.5em}
\end{figure}

This loss function builds on the minimal model of phase change dynamics proposed by \citet{singh2024needs}, where loss is also computed on tensor products of multiple vectors. The tensor product encodes the intuition that multiple sub-circuits need to simultaneously ``go right'' in order for the entire circuit to be learned. The model of \citet{singh2024needs} is equivalent to the loss for a single mechanism in our model. Our full model has \emph{multiple} circuits (\icl and \ciwl) in competition, racing to minimize loss (Section~\ref{sec:whyicl:race}), but also cooperating (Sections~\ref{sec:whyicl:reuse}-\ref{sec:whyicl:iclclose}) via shared sub-circuits ($\mathbf{b}, \mathbf{c}$). 

We model these race dynamics using two mechanisms that interact multiplicatively (lines \eqref{1} and \eqref{2}). If the offset $\mu_1=0$, then minimizing either the loss of Mechanism 1 or 2 will lead to 0 loss.
By setting $\mu_1 > 0$, we can enforce an asymptotic preference for one mechanism over the other. Competition is added via the second term (line \eqref{3}), which drives $\mathbf{a}$ or $\mathbf{d}$ to 0 in order to achieve 0 loss (if $\alpha > 0$). This corresponds to the network pressure we observe for Layer 1 to be part of an \icl or \ciwl strategy. To model cooperative interactions in Layer 2, we re-use vectors $\mathbf{b}, \mathbf{c}$ in both mechanisms. Finally, to model the relative speeds of the two mechanisms, we use different dimensions for $\mathbf{a}$ and $\mathbf{d}$ (larger vectors are slower to learn). 

Putting it all together, we simulate the toy model
dynamics numerically for the setting of $dim(\mathbf{a}) = dim(\mathbf{b}) = dim(\mathbf{c}) = 20, dim(\mathbf{d}) = 160, \mu_1 = 0.1, \alpha=0.1$.
We show results in Fig~\ref{fig:toymodel}a, with additional seeds and details in Appendix~\ref{appx:toymodel}. Our simulation shows a transience of Mechanism 1, where its loss first quickly drops with the initial phase change, but then goes back up, indicating that $\mathbf{a}$ is nearly learned but then eventually goes back to $\mathbf{0}$. This return to $\mathbf{0}$ is caused by the asymptotically preferred $\mathbf{d}$ emerging on a slower timescale and the competition term ($\alpha > 0$). These emergence and subsequent transience dynamics mimic those of \icl, with sub-circuits $\mathbf{b}, \mathbf{c}$ representing the shared Layer 2 ``induction heads.''

Curiously, the toy model dynamics showed a small divot in the formation of Mechanism 2, corresponding to \ciwl (highlighted in orange, Fig~\ref{fig:toymodel}a). Specifically, when Mechanism 1 is nearly fully learned ($\sim5k$ iterations), there's a mild \textit{worsening} of Mechanism 2, before the eventual learning that causes transience of Mechanism 1 and asymptotic dominance of Mechanism 2. 
When we reexamined the loss curves of our actual transformer,  (whose corresponding in-context accuracies are plotted in Fig~\ref{fig:main}b), 
we found a similar divot in the loss on \ciwleval evaluator (Fig~\ref{fig:toymodel}b)! While we do not fully understand what leads to this brief divot,\footnote{We speculate that as Mechanism 1 (ICL) forms, the competition forces a brief regression to Mechanism 2 (CIWL), which is slowly learning in parallel.} we believe this mirroring between model and empirics lends credence to this simple mathematical model.

\begin{figure}
    \centering
    \includegraphics[width=\linewidth]{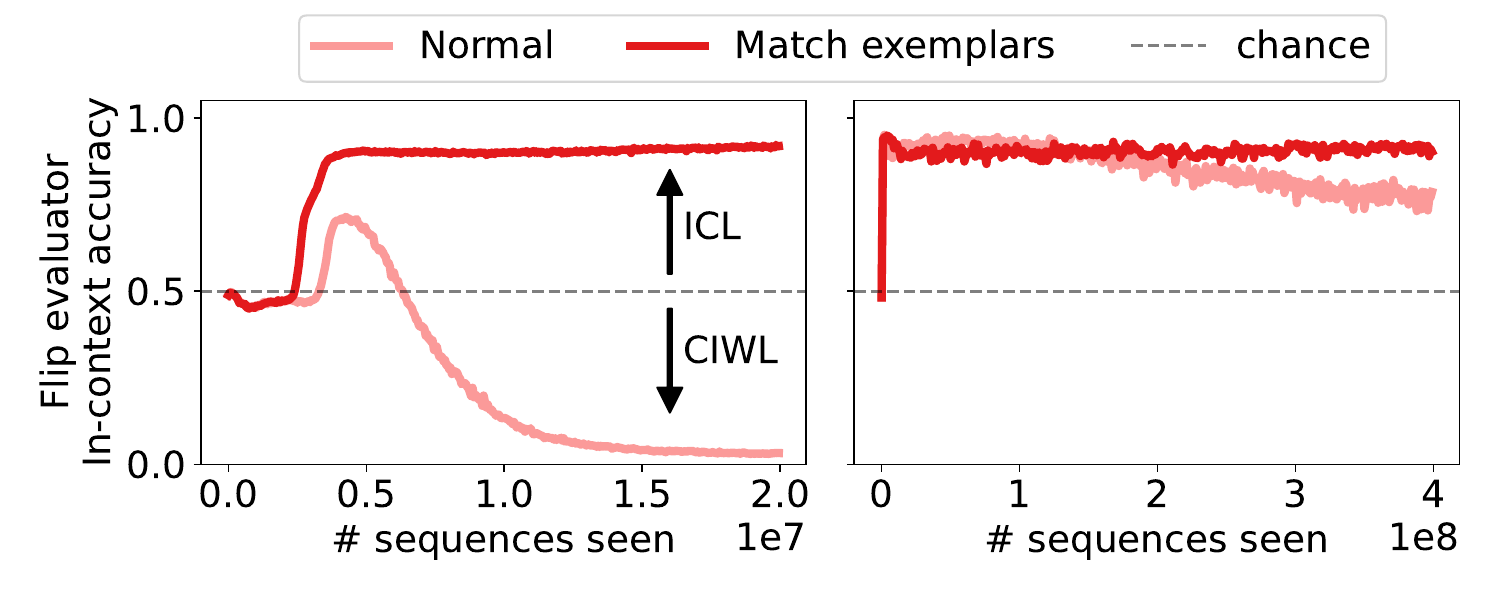}
    \vspace{-3em}
    \caption{Matching context and query exemplars leads to persistent ICL in small (left) and large (right) models, as evidenced by the darker red line converging to accuracy one on the flip evaluator. Larger models are trained with the setup of \citet{singh_transient_2023}. The timescale of transience is much longer in larger models (lighter line, right plot), so we don't see a full fading of ICL. Notably, the matched exemplars run shows no indicating of turning downwards.}
    \label{fig:mq}
    \vspace{-1.5em}
\end{figure}

\section{\icl can be made asymptotically preferred}
\label{sec:mq}

Equipped with a better understanding of transience, we return to a revised version of the question from \cite{Chan2022}: Are there data properties that incentivize the \textit{asymptotic} emergence of \icl, but don't require it? We tackle this question using intuition from our toy model.

Specifically, if $\mu_1=0$, the faster Mechanism 1 (corresponding to ICL) would win asymptotically (Fig~\ref{fig:toymodel0}). Thus, when two strategies are possible, one can be made preferred asymptotically if it's 1) faster and 2) as good asymptotically. 
We've already established that \icl is faster to emerge than \ciwl (Section~\ref{sec:whyicl:race}) when training on our bursty data, so all that remains is to see if we can find data properties that make ICL ``not worse'' at the task than \ciwl.

The key difference in the mechanisms we find for \icl and \ciwl is that \ciwl matches query exemplars to \textit{static labels}, while \icl matches to a \textit{variable context exemplar}. This variation may make the ``matching'' harder for \icl. We ablate this by considering data where the relevant context exemplar is the exact same as the query exemplar.\footnote{Note: there are still 20 exemplars per class, making the exemplar-label mapping many-to-one (leaving \ciwl the same), it's just that the bursty context exemplar exactly matches the query exemplar (making ICL more effective).} In Fig \ref{fig:mq}, we find that training on such data makes \icl asymptotic---it persists after its initial emergence! Furthermore, this finding extends to the larger-scale setup considered by \citet{singh_transient_2023} as well,\footnote{Specifically, 12-layer transformers, with MLPs, trained on bursty sequences of 8 exemplar-label pairs, with the 3 bursty context exemplars matching the query in our intervention.} indicating that the intuitions from the toy model may extend to larger networks.

In combination with \citet{Chan2022}, this result would point toward data properties that not only lead to the emergence of \icl---bursty data, with high diversity (\# of classes) and variation (\# of exemplars) across sequences---but its \textit{persistence} as well: strong correlations in-context---exact matching exemplars. Overall, we found this intervention, motivated by our minimal mathematical model, a compelling illustration of how a better dynamical understanding may lead to data interventions that can control the cycling in and out of strategies throughout training.

\section{Discussion}
When there are multiple ways to solve a problem, when and why does a model ``choose'' between each strategy?

In this work, we have explored this question by building on previous work demonstrating the \emph{transience} of in-context learning (ICL) in transformers---whereby ICL emerges but then disappears after long training times \citep[see extended Related Work in Appendix~\ref{appx:rw}]{singh_transient_2023, anand_dual_2024, he_learning_2024}.

We uncovered a few surprising findings. On a task that was designed to be solvable by both in-weights learning and in-context learning, we find that the model asymptotically prefers an unexpected strategy that is a combination of the two, which we call ``\ciwlfull'' (\ciwl). After first showing this behaviorally through OOD evaluations, we elucidate the mechanistic implementation of this strategy (Section~\ref{sec:ciwl}), which may have connections to notions of task recognition \cite{lin2024dualoperatingmodesincontext} and superposition in larger models \cite{templeton2024scaling}.\footnote{Specifically, the compression of 12800 class-specific skip-trigrams into 8 attention heads with superposition could motivate research on an analog of sparse autoencoders \citet{bricken2023monosemanticity} for attention heads (e.g., going from 8 heads to 12800 sparse heads that may individually represent each skip-trigram feature).} 

We find that \textit{cooperative} interactions between strategies may lead to the transient emergence of one strategy, even though it is asymptotically dominated by the other strategy. In particular, \icl and \ciwl have shared sub-circuitry that enable such cooperation. For the case of transient ICL, we show that its emergence is due to three key properties:\footnote{In addition to our ``positive'' evidence for the found explanation, we reject various alternate hypotheses in Appendix~\ref{appx:rejected}.} \textbf{useful} (ICL helps reduce loss on the task, by construction), \textbf{on the way} (ICL is close to the path of a CIWL solution, Section~\ref{sec:whyicl:iclclose}), and \textbf{fast} (ICL only emerges if CIWL is sufficiently unformed, Section~\ref{sec:whyicl:race}). Eventually, \ciwl does dominate and \icl fades due to competitive interactions, as suggested by prior work \citep{singh_transient_2023}. We borrow the term ``coopetition'' to describe these simultaneously cooperative and competitive strategies.

We crystallize these intuitions from the case study to a more general mathematical model of strategy racing\footnote{\citet{saxe2022neural} also considered the notion of different pathways in a network competing to minimize loss, using the formalism of gated deep linear networks, and found similar intuitions about the ``fastest'' pathway winning.} that captures cooperative and competitive interactions. This toy model motivates experiments that identify data properties that do lead to persistent ICL, despite not requiring it, answering some of the original questions of \citet{Chan2022}.

The shared sub-circuitry we find may also point to different forms of ICL (e.g., task recognition and task learning) not being as different as previously perceived. While earlier works have argued a tradeoff via competitive interactions \cite{lin2024dualoperatingmodesincontext, nguyen_differential_2024, park_competition_2024}, our work suggests a more nuanced take and provides mechanistic evidence for more recent notions that ICL abilities lie on a \textit{spectrum} \cite{lampinen2024broaderspectrumincontextlearning}.

Overall, we hope our work serves to deepen the understanding of ICL and how it may interact with other strategies through the course of training. Beyond ICL, we view our work as a case study demonstrating a few themes that are important to keep in mind, for those of us who wish to understand model behaviors: (1) Models often learn \emph{surprising and counterintuitive} strategies, even for simple tasks. (2) Models are highly \emph{dynamical} and we cannot assume that their strategies remain constant over training, even if they appear stable at a given time. (3) These training dynamics are affected by a kind of \emph{backwards hysteresis}, where later strategies can affect the development of earlier strategies.\footnote{We note that this backwards hysteresis complements notions of path dependence, such as the CIWL strategy being influenced by ICL earlier, for which correlates have been found in larger-scale language models \cite{yin2025attentionheadsmatterincontext}.} (4) Alternate strategies are not always strictly in competition and can exhibit \emph{coopetition} by boosting each other.

\section*{Acknowledgements}
We'd like to thank Andrew Lampinen, DJ Strouse, Kira Düsterwald, Jirko Rubruck, Dan Roberts, Basile Confavreaux, Jin Lee, Yedi Zhang,  and Murray Shanahan for useful discussions and feedback on early drafts.

A.K.S. and T.M. are funded by the Gatsby Charitable foundation. This work was supported by a Schmidt Science Polymath Award to A.M.S., and the Sainsbury Wellcome Centre Core Grant from Wellcome (219627/Z/19/Z) and the Gatsby Charitable Foundation (GAT3850). A.M.S. is a CIFAR Azrieli Global Scholar in the Learning in Machines \& Brains program.

\section*{Dedication}

In loving memory of Felix Hill, whose mentorship, vision, and enthusiasm made this entire research program on in-context learning possible (\citet{Chan2022, singh_transient_2023, singh2024needs}, and now this paper). Felix's ideas continue to influence our field and will resonate for many years to come. More importantly, Felix made an immeasurable and lasting impact on the lives of the researchers he mentored and inspired.

\bibliography{references}
\bibliographystyle{icml2025}

%%%%%%%%%%%%%%%%%%%%%%%%%%%%%%%%%%%%%%%%%%%%%%%%%%%%%%%%%%%%%%%%%%%%%%%%%%%%%%%
%%%%%%%%%%%%%%%%%%%%%%%%%%%%%%%%%%%%%%%%%%%%%%%%%%%%%%%%%%%%%%%%%%%%%%%%%%%%%%%
% APPENDIX
%%%%%%%%%%%%%%%%%%%%%%%%%%%%%%%%%%%%%%%%%%%%%%%%%%%%%%%%%%%%%%%%%%%%%%%%%%%%%%%
%%%%%%%%%%%%%%%%%%%%%%%%%%%%%%%%%%%%%%%%%%%%%%%%%%%%%%%%%%%%%%%%%%%%%%%%%%%%%%%
\newpage
\appendix
\onecolumn

\section{Extended related work}
\label{appx:rw}

\textbf{Different forms of ICL} Since its first documented emergence at scale \cite{GPT3}, few-shot ICL in transformers became widely researched. ICL is often contrasted to in-weights learning, where networks learn tasks over the course of training in-weights, rather than adapting to context. However, some \cite{lin2024dualoperatingmodesincontext} have argued that few-shot learning may be more akin to task recognition, where few-shot prompts serve more to identify a task that the model already knows, rather than learn a new task. This tradeoff may even be modulated by scale \cite{wei2023larger} or training time \cite{wang2024investigatingpretrainingdynamicsincontext}. While induction heads \cite{InductionHeads} may serve as a mechanistic explanation for few-shot task learning, mechanisms for task-recognition-like phenomena have been lacking. In our work, we provide evidence for a context-constrained in-weights mechanism which may help elucidate task-recognition like phenomena by indexing into the label space in context. Furthermore, we hope the shared subcircuitry between different levels of context-usage (\ciwl and \icl) may enhance mechanistic understanding of the full spectrum of in-context learning abilities \cite{lampinen2024broaderspectrumincontextlearning}.

\textbf{A dynamical understanding of ICL} Since the finding that the emergence of ICL may be a \textit{transient} phenomena (\citet{singh_transient_2023}, reproduced by \citet{he_learning_2024, anand_dual_2024} in other settings), many have focused on an enhanced dynamical understanding of ICL. Some have studied emergence dynamics \cite{reddy2023mechanistic, singh2024needs} of ICL, though only on data that requires ICL to minimize loss. Newer work has considered the tradeoff between strategies throughout the course of training in small \cite{nguyen_differential_2024, park_competition_2024} and larger models \cite{wang2024investigatingpretrainingdynamicsincontext}, or centered on theoretical modeling \cite{chan_toward_2024}. These works largely focus on \textit{competitive dynamics} between strategies. Our work goes beyond prior work by providing an explanation for the emergence of ICL \textit{when not asymptotically preferred or necessary} due to \textit{cooperative interactions} between different strategies. 

\section{Extended behaviors seen in small settings}
\label{appx:behavior}
In this section, we provide additional results on alternative setups we considered---specifically, using other types of positional embeddings, interaction with data properties considered by \citet{Chan2022}, and potentially using MLP layers in a 2L network (with absolute positional embeddings). For further exploration of the transience phenomena, we refer readers to the original work of \citet{singh_transient_2023}---our work here focuses on finding a smaller scale setting that allows for study of the key phenomena (with insights that we show may carry to larger settings, Section~\ref{sec:mq}.

All our plots in this section feature accuracy on the IWL evaluator as well. Note that in-context accuracy on this evaluator is meaningless, as the correct label does not appear in context. As a result, we just plot the direct accuracy (for which chance level would be $\frac{1}{\# classes}$). Most of the times, this accuracy stays at chance level, though we note settings where it seems to rise. Further exploration of mechanisms that may be responsible IWL are provided in \ref{appx:results:iwl}.

Runs in this section are also often run for longer than our main experiments---we truncated the plots in the main paper to focus on the relevant phenomena. Here, we present the full timecourses we ran in case it's useful for future researchers. Code will also be open-sourced upon acceptance. % TODO update

\subsection{Different random seeds}
\label{appx:behavior:seeds}

\begin{figure}
    \centering
    \includegraphics[width=0.8\linewidth]{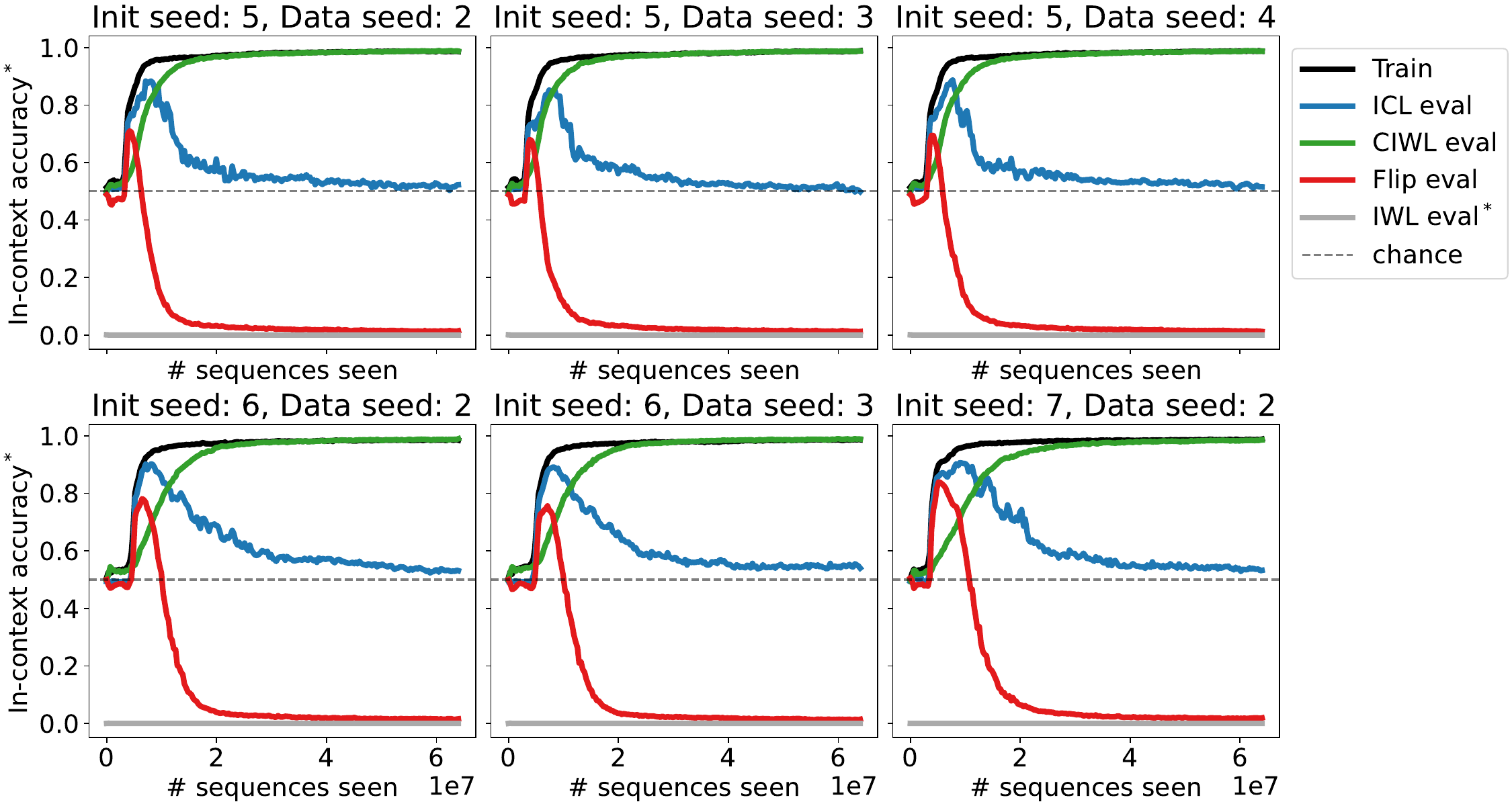}
    \caption{Reproduction of our main setup (Section~\ref{sec:setup}) over random seeds. Top left figure is same as Figure~\ref{fig:main}b.}
    \label{fig:seeds}
\end{figure}

To confirm our reproduction, we experimented over a few seeds (for both model initialization and data generation/ordering)---see Figure~\ref{fig:seeds}. We found data seed to make little-to-no difference. Model intialization can change the exact timing of transience, but the general profiles of all the curves are the same.

\subsection{Other positional embeddings}
\label{appx:behavior:posemb}

We considered absolute sinusoidal positional embeddings (introduced by \citet{vaswani2017attention} and used in the original works of \citet{Chan2022} and \citet{singh_transient_2023}) as well as rotary positional embeddings (RoPE, introduced in \cite{rope} and used by \cite{singh2024needs}). However, we found that, for two-layer attention only networks, learned absolute positional embeddings were needed to reproduce the emergence and transience of ICL. While larger networks don't require this added flexibility, we found it necessary to elicit emergent ICL, which we speculate may be due to the difficulty of learning previous token heads in Layer 1 otherwise \cite{InductionHeads}.

\begin{figure}
    \centering
    \includegraphics[width=0.8\linewidth]{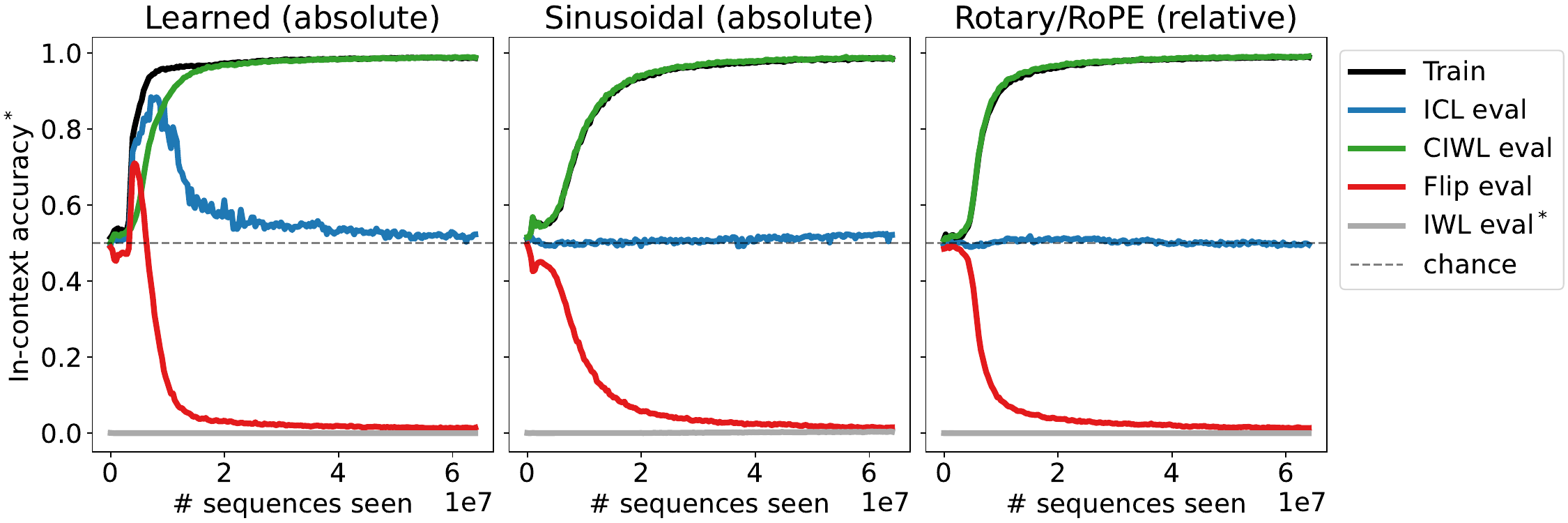}
    \caption{Changes in behavior when using different positional embedding schemes. Left figure is the same as Figure~\ref{fig:main}b. Note that learned absolute positional embeddings are needed to get ICL emergence in a 2L attention-only transformer on our task.}
    \label{fig:posemb}
\end{figure}

\subsection{ICL does not emerge with lower classes or fewer exemplars}
\label{appx:behavior:data}

\begin{figure}
    \centering
    \includegraphics[width=0.8\linewidth]{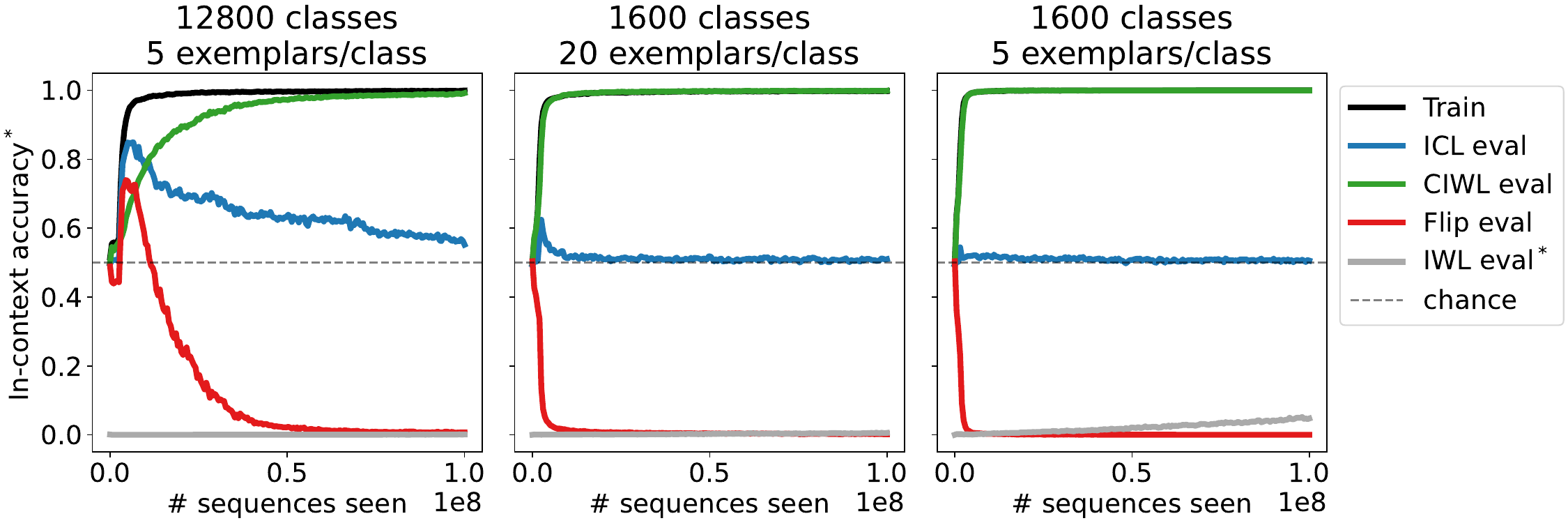}
    \caption{Changes in behavior under different data properties, largely reproducing what was seen by \citet{singh_transient_2023}.}
    \label{fig:dataprop}
\end{figure}

We consider different \#'s of classes and different \#'s of exemplars. \citet{Chan2022} found that higher numbers of both incentivized ICL emergence, and \citet{singh_transient_2023} found that more classes slowed down transience. In Figure~\ref{fig:dataprop}, we find that larger number of classes are necessary for \icl to dominate over \ciwl for a time (Flip evaluator accuracy $> 0.5$). We still see smaller transience effects with 1600 classes, as long as 20 exemplars are used. When reducing both (1600 classes, 5 exemplars), we see no \icl emergence, and even some asymptotic strengthening of pure in-weights mechanisms. 

These findings match what was seen by \citet{singh_transient_2023}, and we hypothesize that when \# of classes is lower, the \ciwl mechanism emerges ``too quickly'' for \icl to show up (see Section~\ref{sec:whyicl:race}).

\subsection{Behavior with MLPs}
\label{appx:behavior:mlp}

\begin{figure}
    \centering
    \includegraphics[width=\linewidth]{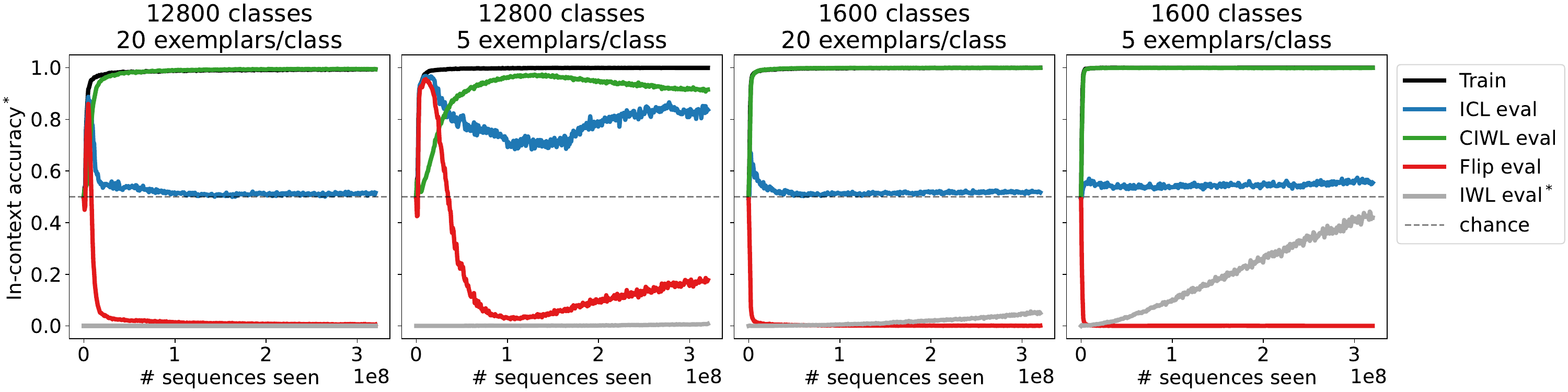}
    \caption{Changes in behavior when training 2-layer transformers with MLPs. In our main setting (left), we see similar trends, but when training on 12800 classes with 5 exemplars each, we get transient and resurgent ICL (unlike the corresponding run without MLPs, Figure~\ref{fig:dataprop}, left) In the rightmost plot, we see significantly increased pure in-weights learning at low numbers of classes and exemplars (as compared to Figure~\ref{fig:dataprop}), which may be due to the hypothesized role of MLPs in such pathways \cite{geva2021transformer, meng2023locating, singh_transient_2023}.}
    \label{fig:mlp}
\end{figure}

We also considered the use of 2-layer transformers with MLPs as well, as are typically used in practice \cite{vaswani2017attention}. We use a standard 4x expansion factor and GeLU activations \cite{hendrycks2023gaussianerrorlinearunits}. Results in Figure~\ref{fig:mlp}.

While our main setting (12800 classes, 20 exemplars) seems to have similar trends, we observed transience and resurgence in other settings (12800 classes, 5 exemplars), which make it difficulty to claim asymptotic behaviors in these small models with MLP. We thus restricted our main analysis to the attention-only models, as is common for mechanistic work on transformers \cite{anthropicMathFramework, InductionHeads, singh2024needs}.

\section{Additional results}
\label{appx:results}

\begin{figure}
    \centering
    \includegraphics[width=0.8\linewidth]{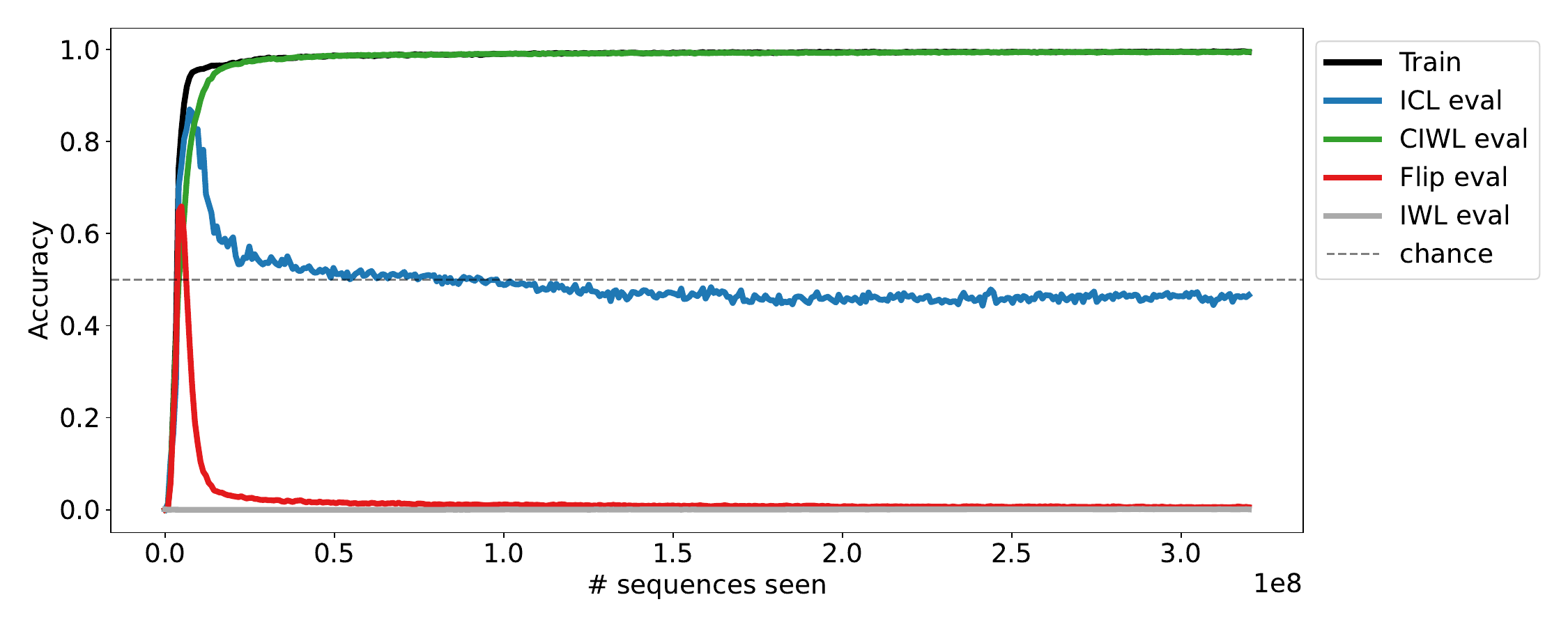}
    \caption{Same run as Figure~\ref{fig:main}b, but run 16x as long. We plot accuracy on the y-axis here (across all 12800 labels) as opposed to in-context accuracy to more compellingly demonstrate saturation of \ciwl (green) and minimal learning of IWL (gray). This plot lets us include that networks are in fact not learning a pure IWL mechanism, but rather a context-constrained one which requires the correct label in context.}
    \label{fig:extended-main}
\end{figure}

\begin{figure}
    \centering
    \includegraphics[width=0.3\linewidth]{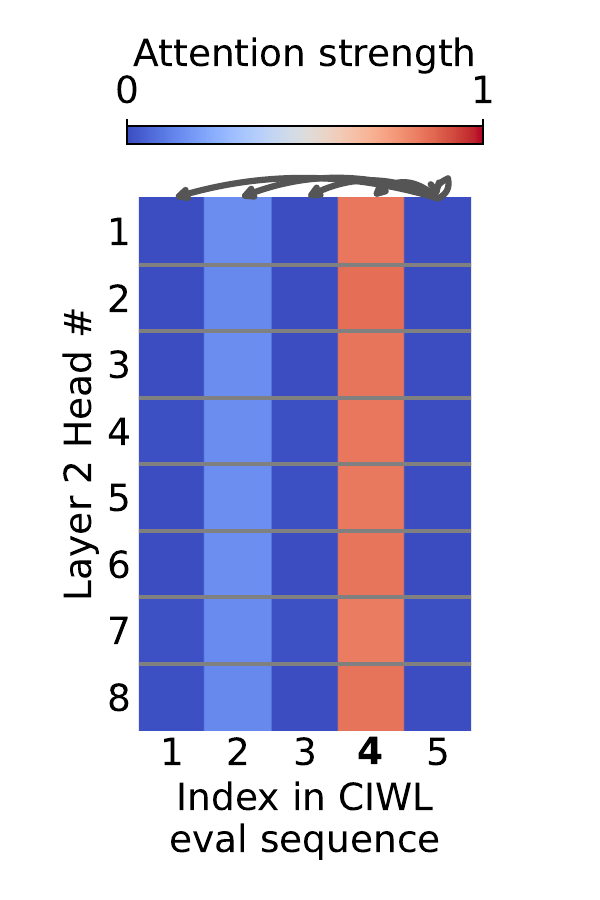}
    \caption{Analogous plot to Figure~\ref{fig:ciwl}a, except the label is inserted at index 4 as opposed to index 2.}
    \label{fig:ciwl2}
\end{figure}

\begin{figure}
    \centering
    \includegraphics[width=\linewidth]{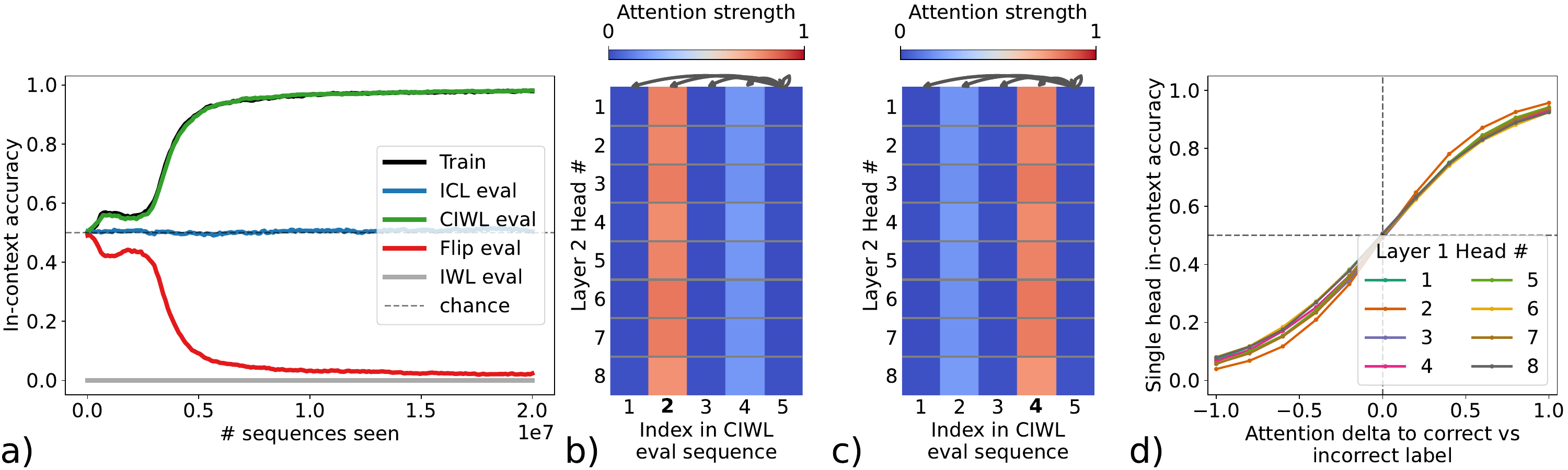}
    \caption{A 1-layer transformer trained on our bursty training data can learn skip-trigram-copiers, just as well as the 2-layer model. \textbf{(a)} Metrics indicate that the model learns a \ciwl strategy. Notably, \icl does not emerge, as we would expect for a 1-layer transformer \cite{anthropicMathFramework}. \textbf{(b-c)} Analogous plots to Figure~\ref{fig:ciwl}a and Figure~\ref{fig:ciwl2} showing that all heads learn attention from query token (index 5) to the correct token in-context (index 2 or 4 for b, c respectively). \textbf{(d)} Analogous plot to Figure~\ref{fig:ciwl}b showing that all heads are copiers.}
    \label{fig:ciwl1L}
\end{figure}

\begin{figure}
    \centering
    \includegraphics[width=\linewidth]{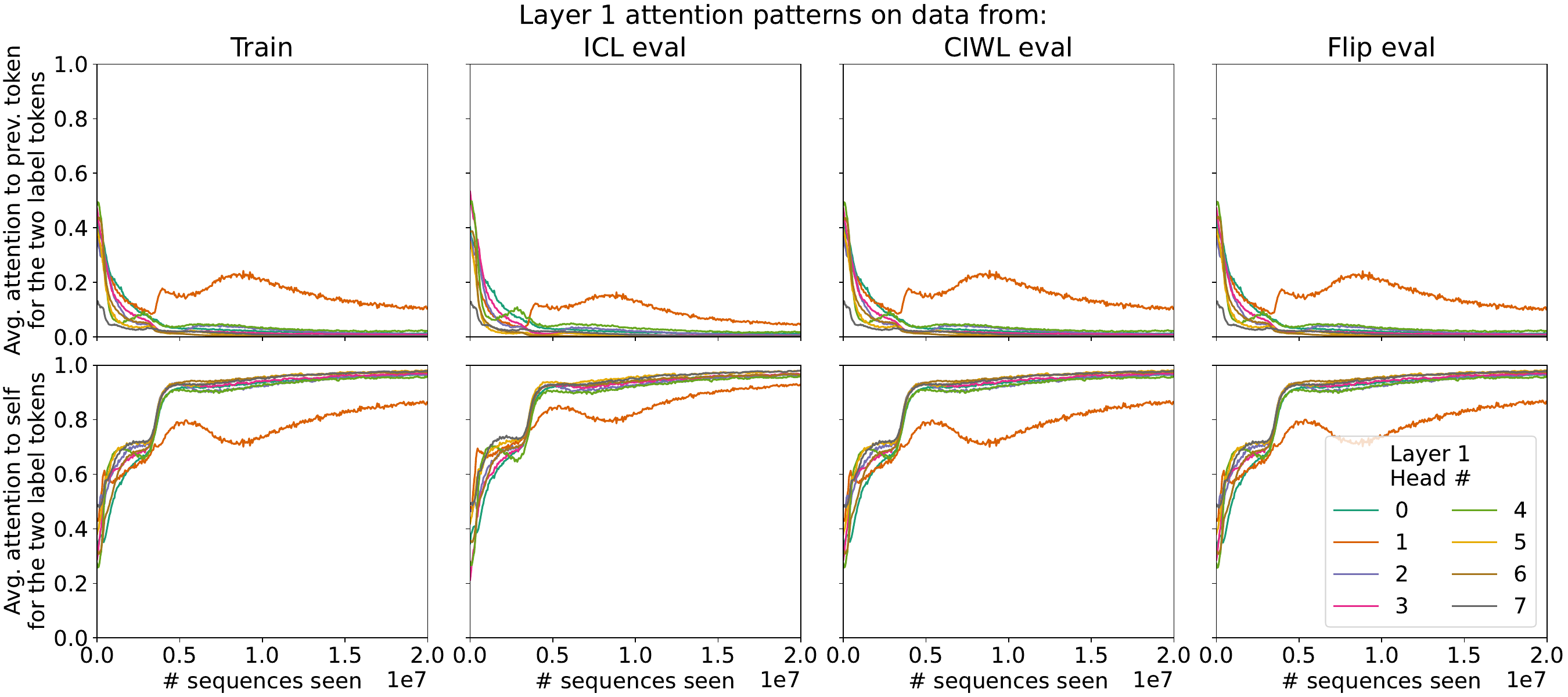}
    \caption{Per-Layer-1-head average attention patterns over different data subsets through the course of training. Top row depicts average attention from a label token (at index 2,44) to the previous token (index 1, 3, respectively). Bottom row depicts average attention from a label token (at index 2, 4) to itself. Most of the attention on these tokens is made up of these two pieces (attention to previous and attention to self), though tokens can attend to all previous tokens (standard causal transformer setup). Attention patterns look similar on similar subsets, indicating that Layer 1 heads operate similarly on our out of distribution evaluators. Layer 1 Head 1 appears to have transient previous token behavior, suggesting it may be partly responsible for the emergence of ICL. In Appendix~\ref{appx:}}
    \label{fig:l1attn}
\end{figure}

\subsection{Networks barely exhibit \textit{pure} in-weights learning (IWL)}
\label{appx:results:iwl}

Figure~\ref{fig:extended-main} depicts accuracy on a pure IWL evaluator (Figure~\ref{fig:main}a, gray), which differs from the \ciwleval evaluator in that the correct label token does not appear in context. This evaluator is the same as that used by \citet{Chan2022, singh_transient_2023}. Like \cite{singh_transient_2023}, we find that networks perform relatively poorly on thie evaluator, instead employing a \ciwl strategy (as evidenced by the delta between green and gray curves in Figure~\ref{fig:extended-main}).

That said, the performance on the IWL evaluator at the end of Figure~\ref{fig:extended-main} is 0.08\%, which is above chance level (which would be 1/12800 $\approx 0.008\%$, indicating that the network may have picked up a tiny amount of pure IWL (or it could be noise). This prompted us to investigate further, and we did uncover a very minor in-weights mechanism.

Specifically, if we ablate all attention heads (so that the network is just the embedding $\rightarrow$ unembedding connection), we find that in-context accuracy on \ciwl is 62.46\% (compared to 98.74\% with no ablation). Note, with this no-heads ablation, the network can't actually attend to any tokens in context, so the ``in-context accuracy'' can be interpreted as ``how likely is the model to output the correct class over a given random class.'' This value (62.46\%) being above chance (50\%) indicates the model is able to do this comparison somewhat accurately using just the embedding $\rightarrow$ unembedding pathway.

To make sure this pure IWL mechanism isn't playing a key role in our results, we consider the opposite ablation: zero-out the embedding $\rightarrow$ unembedding pathway, and leave the rest of the network unchanged. We do this by using a pattern-and-value preserving ablation, using two passes through the network. In the first pass, no intervention is used, and patterns and values are cached. In the second pass, we 0 out the embeddings, but use the cached activations for the attention heads, having the intended effect. This ablation leads to 2\% drop in accuracy (98.74\% to 96.72\%), indicating that if the \ciwl mechanism, implemented via skip-trigram-copying attention heads, is activate, the embedding $\rightarrow$ unembedding pathway is quite auxiliary.

To conclude, networks do a small amount of \textit{pure} IWL, even when trained on ``bursty data,'' but the contribution of this strategy is negligible compared to the more dominant \icl and \ciwl trading off through training. 

\begin{figure}
    \centering
    \includegraphics[width=\linewidth]{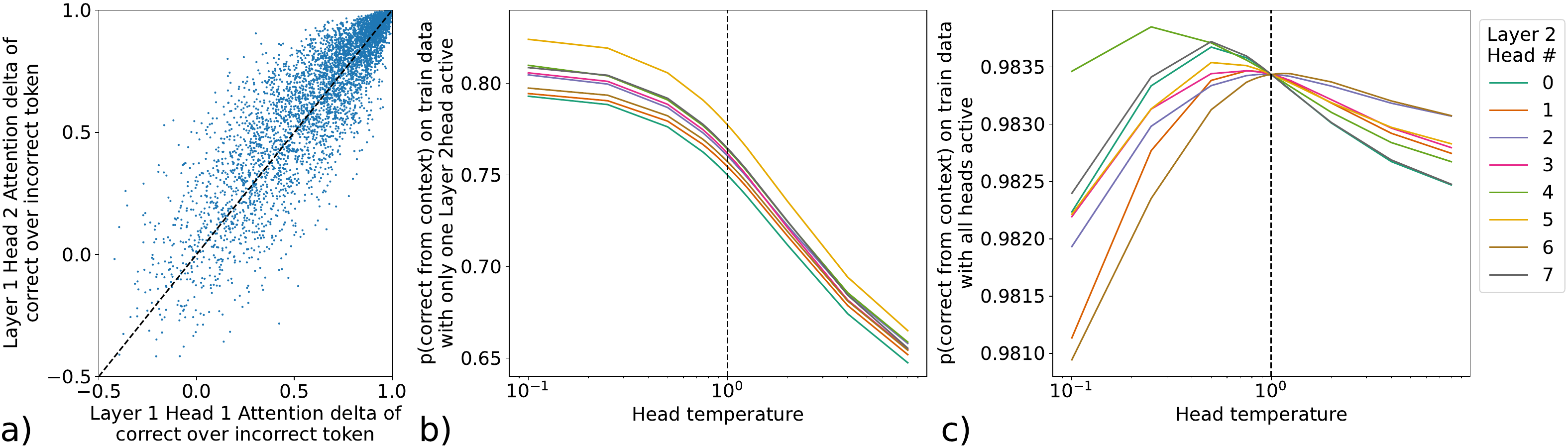}
    \caption{Preliminary evidence of superposition. \textbf{(a)} While average attention patterns across heads look identical (Figure~\ref{fig:ciwl}a), patterns on individual points can be quite different (e.g., shown for Head 1 and 2, where each point in the plot represents attention deltas of these two heads for a single evaluation sequence). \textbf{(b)} Decreasing the temperature of any Layer 2 head's attention operation, if it is the only Layer 2 head active, increases performance. \textbf{(c)} The same temperature decrease does not as consistently affect performance if all heads are active, indicating non-additive interactions.}
    \label{fig:superposition}
\end{figure}

\subsection{Layer 2 heads may be acting in superposition}
\label{appx:results:superposition}

In Section~\ref{sec:ciwl}, we identify that the asymptotic strategy of the network relies on Layer 2 heads acting as skip-trigram-copiers and thus implementing a \ciwl strategy. However, we observe some curious features:
\begin{enumerate}
    \item All heads appear to be doing the same thing on average (Figure~\ref{fig:ciwl}a).
    \item All heads are \textit{imperfect} on average---attention to the correct token never converges to 1 (instead plateauing around 0.8).
\end{enumerate}

In this section, we lay out preliminary evidence for how these heads may be acting in \textit{superposition}. Our hypothesis was inspired by \citet{elhage2022superposition}, who find that transformer MLP hidden layers may represent sparse features in superposition. The skip trigrams our network is learning are sparse (each sequence only requires comparing 2), and numerous (12800 classes), while the network weights that can be used to represent these are limited (eight 8-dimensional heads).

First, we find that while heads have the same attention patterns on average, their patterns on a given sequence can be quite different (Figure~\ref{fig:superposition}a). This would point to some form of distributed computation (which is hidden if only considering average patterns).

The most compelling evidence we find is by considering the interactions between different heads, and their imperfect average attention. If we consider a head in isolation (with all the other heads ablated), we can ask if the imperfection is ``optimal.'' Specifically, we consider artificially decreasing the temperature of the softmax attention on this head,\footnote{Note when conducting these experiments, we artificially also only allow the head to attend to label tokens. We know this is what they do in practice, and we wanted to avoid high temperature performance from being artificially hurt by increased attention to non-label tokens.} to see if its performance\footnote{By performance here, we mean performance on the task if only this head is active. Alternatively, this value can be viewed as a readout of the contribution of this head to the output..} is better or worse. We see in Figure~\ref{fig:superposition}b that uniformly decreasing the temperature improves performance, which begs the question: Why doesn't the network learn to do this?

The answer comes from considering how heads interact. If we leave all heads active and then decrease the temperature of a head, we find that overall performance often \textit{goes down}. Each head on its own gets better at solving the task, but the inter-head interactions can often make the network overall worse! These non-additive interactions hint at a tight coupling between heads, and the notion that heads are meaningfully exploiting their dynamic range of attention (instead of the simplified, binary notion of ``heads learn to attend to the correct token''). These results are in stark contrast to the additivity of heads found by \citet{singh2024needs}, and we suggest this is due to a form of superposition.

We conclude by noting that this suggestion is somewhat speculative, backed by preliminary evidence. We hope this heads-in-superposition hypothesis, and preliminary evidence supporting it, can motivate further work on understanding individual attention head function.

\begin{figure}
    \centering
    \includegraphics[width=\linewidth]{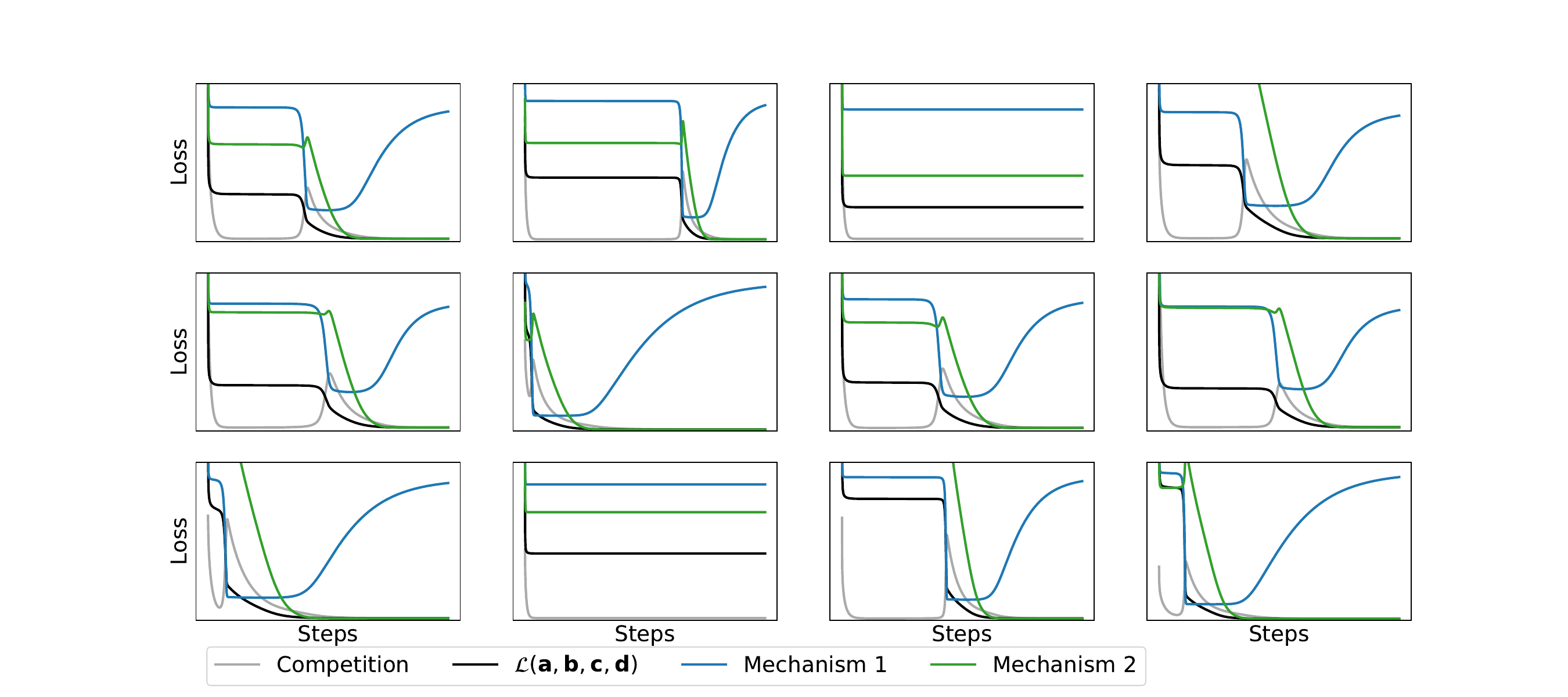}
    \caption{Simulations of the toy model over 12 seeds, using the same settings as those for Figure~\ref{fig:toymodel}: $dim(\mathbf{a}) = dim(\mathbf{b}) = dim(\mathbf{c}) = 20, dim(\mathbf{d}) = 160, \mu_1 = 0.1, \alpha=0.1$. We additionally show the competition term in gray.}
    \label{fig:toymodelseeds}
\end{figure}

\begin{figure}
    \centering
    \includegraphics[width=\linewidth]{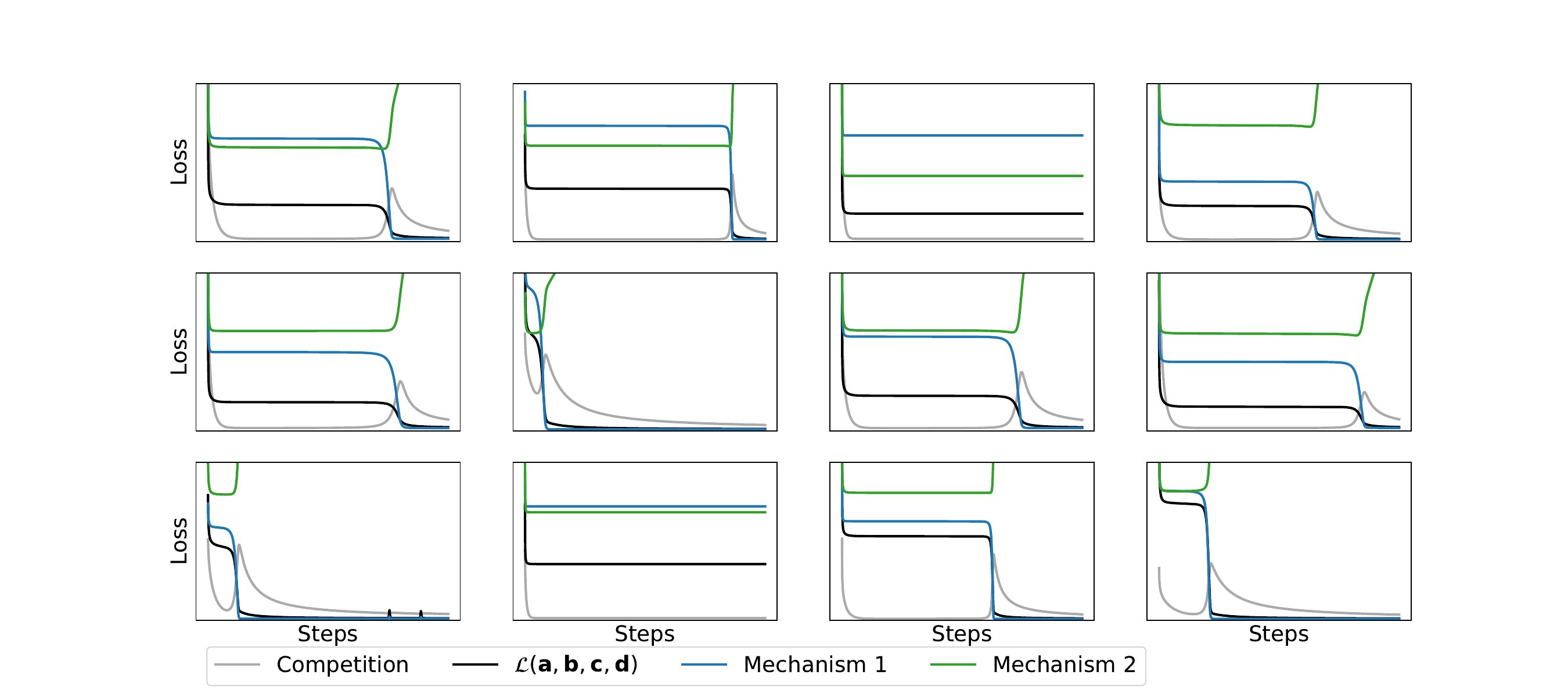}
    \caption{Simulations of the toy model over 12 seeds, using the settings: $dim(\mathbf{a}) = dim(\mathbf{b}) = dim(\mathbf{c}) = 20, dim(\mathbf{d}) = 160, \mu_1 = 0, \alpha=0.1$. Without the asymptotic bias $\mu_1 = 0$, the faster Mechanism 1 gets learned first and dominates. $\mathbf{d} \rightarrow \mathbf{0}$ due to the competition term, which leads to increasing loss of Mechanism 2 (green).} 
    \label{fig:toymodel0}
\end{figure}

\section{Additional details and seeds on toy model}
\label{appx:toymodel}

All vectors in our toy model (Section~\ref{sec:toymodel}) are initialized randomly with each element being drawn from $\mathcal{N}(0,1)$. Vectors evolve through gradient descent, with a learning rate of 1. In practice, we use a scaled Frobenius norm in the loss function, where we normalize for the size of the tensors in each expression, to avoid needing to tune the learning rate. Mathematically, this rescaling is isomorphic to a corresponding setting of $\mu_1, \alpha$ and learning rate.

In Figure~\ref{fig:toymodelseeds}, we show simulations on 12 random seeds (to supplement the single seed shown in Figure~\ref{fig:toymodel}). While the exact timing in magnitudes of the curves can shift around, the overall dynamical profile (transience of Mechanism 1, the divot in Mechanism 2 formation) is largely preserved. For some seeds, we found that the vectors did not exit the loss plateau (likely caused by a saddle point in the loss landscape, c.f. \cite{singh2024needs}) for the duration of our simulation.

In Figure~\ref{fig:toymodel0}, we show simulations on 12 random seeds with the same settings as Figure~\ref{fig:toymodelseeds}, except that $\mu_1 = 0$ instead of $\mu_1 = 0.1$. Eliminating the asymptotic bias for Mechanism 2 gets rid of transient behavior, with the faster mechanism (Mechanism 1) emerging and dominating. These intuitions motivated our experiments in Section~\ref{sec:mq}.

All code will be open-sourced on publication, to allow researchers to easily play around with this model and build further intuitions. We're especially excited to see if this toy model can model other types of dynamics (e.g., recent work from \citet{zhang2025trainingdynamicsincontextlearning} mentions similar interactions in their discussion).

\section{Rejected hypotheses for the transience of emergent \icl}
\label{appx:rejected}

In the scientific process of figuring out what may cause ICL emergence, given that it's not asymptotically preferred, we considered (and subsequently rejected) many alternative hypotheses. In the spirit of sharing the path used to arrive at a conclusion, we share some of these hypotheses here.

\begin{figure}
    \centering
    \includegraphics[width=0.8\linewidth]{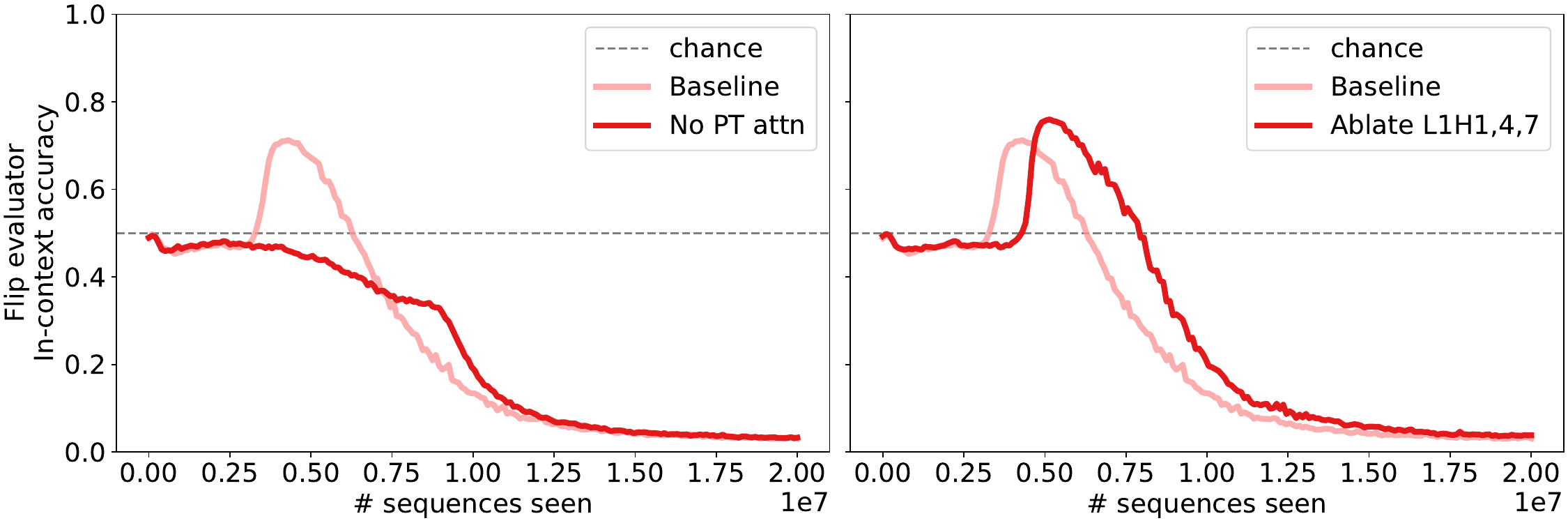}
    \caption{Various experiments to reject alternative hypotheses for why \icl emerges, despite not being asymptotic. Recall that accuracy of 1 on the Flip evaluator means \icl is dominant, and accuracy of 0 means \ciwl is dominant. \textbf{(a)} \icl is not necessary for \ciwl emergence. Darker line indicates evolution from a training run where Layer 1 heads were prevented from attending to the previous token, effectively disabling \icl. Lighter line is from standard training (same as Figure~\ref{fig:main}b). This clamp has relatively little effect on the dynamics of \ciwl emergence. \textbf{(b)} \icl is likely not an initialization artifact. Darker line indicates evolution from a training run where Layer 1 heads 1, 4, and 7 (which play a crucial role in \icl in standard training) are ablated. While \icl emergence is slightly delayed, it still occurs, and in fact transience is a bit slower as well.}
    \label{fig:rejected}
\end{figure}

\subsection{Hypothesis 1: Earlier \icl is \textit{necessary} for \ciwl to emerge}
\label{appx:rejected:iclnotneeded}

Given the experiments in the main paper, this hypothesis already seems a bit suspicious---in Figure~\ref{fig:ciwl1L} and Section~\ref{sec:whyicl:race} we show that \ciwl can be learned without earlier ICL emergence. But what if somehow bursty data requires \icl to emerge first in order to reach a \ciwl solution?

We test (and reject) this hypothesis by considering an experiment where we prevent Layer 1 heads from attending to the previous token, and then training on bursty data. This experiment is a causal intervention on dynamics \cite{singh2024needs}, as it prevents \icl emergence but lets other strategies develop naturally. We show results in Figure~\ref{fig:rejected}a. Notably, we get similar curves to standard training, with the exception that \icl just doesn't emerge (since it was blocked). Thus, we reject the hypothesis that transient \icl is \textit{necessary} to arrive at asymptotic \ciwl.

\subsection{Hypothesis 2: \icl emerges due to a lottery ticket}
\label{appx:rejected:icllottery}

Another idea may be that \icl simply emerges since typical transformer initializations just happen to be ``close'' to an \icl solution---so rather than the dynamics leading the model through an \icl solution transiently, it just starts near one and the transience we observe is more of an ``initialization effect.'' Though not quite the same, the notion of lottery tickets \cite{lottery_ticket} is what inspired this hypothesis.

Given our earlier experiments, this hypothesis does feel a bit suspicious as well---for example, we find that \icl transience is quite robust over initialization seeds (Appendix~\ref{appx:behavior:seeds}). To offer more evidence against this hypothesis, we again consider a more causal intervention on dynamics \cite{singh2024needs}:

If \icl is transient due to initialization effects, this would imply that the circuits being used in \icl are already ``largely present'' at the start of training. Through ablations on checkpoints from standard training, we identify that (for the main run we analyze throughout the text and presented in Figure~\ref{fig:main}b) the Layer 1 heads contributing previous-token behavior (Figure~\ref{fig:l1attn}) to ICL are heads 1, 4, and 7. We establish this by first inspecting attention patterns, and then ablating these heads in checkpoints from around the peak performance of \icl ($\sim$3e6 to $\sim$1.5e7). Such an ablation on these checkpoints eliminates \icl behavior (as verified by performance on the \icleval evaluator dropping to chance level).

We then re-train the model from scratch, with the same initialization, but with these Layer 1 heads clamped off \textit{throughout training}. Presumably, this would eliminate any ``already-existing-at-initialization'' \icl circuits. Doing so minorly delays \icl emergence (Figure~\ref{fig:rejected}b), but does not prevent it. This indicates that their is strong dynamical pressure for \icl to emerge, regardless of initialization, providing evidence against this hypothesis.

\end{document}